\documentclass[twocolumn]{article}
\usepackage{preprint}
\usepackage{hyperref}
\usepackage[authoryear,round]{natbib}

\hypersetup{
    colorlinks=true,
    linkcolor=blue,
    urlcolor=blue,
    citecolor=blue,
    anchorcolor=black
}
\usepackage[utf8]{inputenc}	
\usepackage[T1]{fontenc}
\usepackage[table,dvipsnames]{xcolor}
\usepackage{lineno}					
\usepackage{tikz} 					
\usetikzlibrary{positioning, arrows.meta, shadows, fit, backgrounds}
\usepackage{fancyhdr}
\usepackage{geometry}
\usepackage{doi}
\usepackage{amssymb}
\usepackage{amsmath}
\usepackage{bbm}
\usepackage{bm}
\usepackage{array,booktabs}
\usepackage{appendix}
\usepackage{authblk}
\usepackage{placeins}
\usepackage{comment}
\usepackage{siunitx}
\usepackage{float}
\usepackage{url}
\usepackage{tabularx}
\usepackage{caption}
\usepackage{subcaption}
\usepackage{makecell}
\usepackage{textcomp}
\usepackage{gensymb}
\usepackage{threeparttable}
\usepackage{cleveref}
\usepackage[euler]{textgreek}
\usepackage{enumitem}
\newcolumntype{Y}{>{\centering\arraybackslash}X}
\usepackage{manyfoot}%
\usepackage{algorithm}%
\usepackage{algorithmicx}%

\usepackage{listings}%

\usepackage{titlesec}
\titlespacing\section{0pt}{12pt plus 3pt minus 3pt}{1pt plus 1pt minus 1pt}
\titlespacing\subsection{0pt}{10pt plus 3pt minus 3pt}{1pt plus 1pt minus 1pt}
\titlespacing\subsubsection{0pt}{8pt plus 3pt minus 3pt}{1pt plus 1pt minus 1pt}
\makeatletter
\def\@fnsymbol#1{\ensuremath{\ifcase#1\or *\or \dagger\or \ddagger\or
   \mathsection\or \mathparagraph\or \|\or **\or \dagger\dagger
   \or \ddagger\ddagger \else\@ctrerr\fi}}
\makeatother

\title{Effects of Laser Beam Width and Measurement Noise on Tree Diameter Estimation in Laser Scanning}

\usepackage{authblk}

\author[1\small\ensuremath{\ast}]{Jesse Muhojoki}
\author[1]{Nuutti Ervasti} 
\author[1]{Eric Hyypp\"{a}} 
\author[2]{Aapo Erkkil\"{a}} 
\author[2]{Eppu Honkanen}
\author[2]{Tuomas Yrttimaa} 
\author[1,2]{Mikko Vastaranta} 
\author[1]{Juha Hyypp\"a }

\affil[1]{Department of Remote Sensing and Photogrammetry, Finnish Geospatial Research Institute FGI, The National Land Survey of Finland, Vuorimiehentie 5, 02150 Espoo, Finland}
\affil[2]{School of Forest Sciences, University of Eastern Finland, 80101 Joensuu, Finland}

\begin{document}

\twocolumn[\begin{@twocolumnfalse}

\maketitle

\begin{abstract}

Laser scanning has been utilized for decades to measure tree structure and geometry. Despite continuous advances in sensor technology, uncertainties arising from scanner characteristics and measurement conditions has remained relevant sources of error. In particular, factors such as laser beam diameter can introduce both systematic and random errors in the reconstructed tree geometry. This study investigates the main sources of these errors in mobile laser scanning (MLS) and quantifies their effects on stem curve (stem diameter) estimation. We discovered through simulations and experiments that the heteroscedasticity (difference in variance in ranging and scan angle) in the measurement noise is a key factor along with the beam width of the lidar, which cause the bias of the diameter estimates to depend not only on distance, but also on the tree diameter. We demonstrated that this bias can be reliably predicted and negated for trees with a diameter at the breast height greater than \qty{10}{\centi\metre} using technical specifications of the sensor without any additional calibration. In experiments, the proposed bias compensation method reduced the root-mean-square error of stem diameter estimates to sub-centimeter level on boreal forest test sites.  However, for small trees with diameter below 10 cm, the diameter bias becomes sensitive to the diameter of the tree---an unknown factor---and other factors, such as different bark and out-of-roundness of the stem, limiting the precision of lidar for measuring geometry of trees and applicability of the presented method. 
\end{abstract}

 \vspace{0.3cm}
\vfill\hrule\vspace{2pt}
\enlargethispage{2\baselineskip}
\footnotesize This manuscript version is made available under the CC-BY-NC-ND 4.0 license \url{https://creativecommons.org/licenses/by-nc-nd/4.0/}.
\\

\vspace{0.5cm}

\end{@twocolumnfalse}]

\renewcommand{\thefootnote}{\small\ensuremath{\ast}}
\footnotetext[1]{Corresponding author: Jesse Muhojoki (jesse.muhojoki@nls.fi)}
\renewcommand{\thefootnote}{\arabic{footnote}}
\setcounter{footnote}{0}



\section{Introduction}
\label{sec: intro}

Inventories are a key to providing accurate and up-to-date information of forests at country-level. This information can be used for optimizing the forest usage between harvesting, conservation and recreational use.

Airborne laser scanning (ALS) enables efficient, large-area forest inventories, including individual-tree-level inventories\citep{hyyppa_detecting_1999, persson_detecting_2002, ma_individual_2023, hyyppa_concepts_2024, white_enhanced_2025}. However, due to high flight altitude required for fast collection of the data and the canopy obstructions, the tree stems get few laser hits, and thus, their geometries are not directly measurable from the point clouds. Therefore, reference data are required for the prediction of tree stem attributes from ALS data. In addition to the accuracy requirement of the reference, it needs to be quick to collect and update as trees are growing and there are also disturbances abruptly changing forest structure. Furthermore, the forest change is accelerated by the climate change \citep{andersonteixeira_altered_2013, thom_will_2022, bratu_resilience_2025}. One promising alternative to manual reference data collection is mobile laser scanning (MLS) \citep{liang_use_2014, pires_mobile_2024, holvoet_mobile_2025, liikonen_measuring_2026}. MLS utilizes a lidar and an inertial measurement unit (IMU), potentially augmented with a global navigation satellite system (GNSS) receiver, for acquiring a point cloud while moving for efficient high-coverage data collection in contrast to stationary terrestrial laser scanning (TLS). The tree geometry is typically estimated from the laser scanning point clouds through geometric modeling such as circle or cylinder fitting. Acquisition of MLS data from forest sample plots is significantly faster than measurements with calipers and clinometers. Furthermore, MLS data can be used to describe the entire stem curve, i.e., the stem diameter as a function height---as opposed to just the diameter at breast height---and can provide it with better than \qty{10}{\percent} accuracy \citep{hyyppa_under-canopy_2020, stovall_comparing_2023, muhojoki_benchmarking_2024}. Stem curve enables direct estimation of the stem volume, and the geometric information it provides can be used for estimation of the length and volume of the sawlog and pulpwood. Similar accuracy has been achieved for diameter at breast height (DBH) measurements by \citep{ryding_assessing_2015, bauwens_forest_2016, cabo_comparing_2018, kuzelka_inventory_2022, hoffren_evaluation_2024, vandendaele_mobile_2022}. It should be noted that the accuracy of the stem attribute estimation depends on tree species \citep{yrttimaa_exploring_2022, holvoet_mobile_2025, tavi_combining_2026} and forest structure \citep{ryding_assessing_2015, bauwens_forest_2016, cabo_comparing_2018}. Due to the capacity to record more structural information on individual trees, MLS can also enhance forest inventories by providing alternatives for assessing carbon storage \citep{coops_advances_2025} and vigor \citep{gerbrecht_estimation_2025} of each trees as well as biodiversity indicators, such as structural complexity and the amount of dead wood from each measured sample plot \citep{fol_evaluating_2023}.

In forest inventories, it is essential that measurements of trees and sample plots are unbiased, so that errors do not accumulate when producing estimates over large areas \citep{kangas_forest_2006}. Although MLS systems enable rapid measurement of sample plots and trees, the systematic and random errors inherent in such measurements and their underlying causes remain poorly understood. The tree diameter estimated from laser scanning point clouds through circle or cylinder fitting has been reported to lead to both underestimating \citep{gollob_forest_2020, di_stefano_mobile_2021, kuzelka_inventory_2022, hoffren_evaluation_2024} and overestimating the diameter \citep{ringdahl_enhanced_2013, forsman_bias_2018, muhojoki_benchmarking_2024, balestra_advancing_2024}. It is known that the finite beam width of the laser pulses cause part of this bias \citep{forsman_bias_2018, muhojoki_benchmarking_2024, faitli_integration_2024}, but only few attempts have been made to study this or correct the effects.
\citet{ringdahl_enhanced_2013} proposed a geometric correction method that moves the edge points of the measured circle sector towards the center of the fitted circle, depending on the viewing angle and beam width. While the method reduced the overestimation of the tree diameter from \qty{40}{\percent} to \qty{12}{\percent}, the method used a simplistic model of a laser distance measurement, and is therefore limited to laser scanners with a large beam divergence that greatly overestimates the stem diameter.

\citet{forsman_bias_2018} performed a simulation study and concluded that tree diameter bias increased quadratically as a function of measurement distance. However, studies by \citet{kuzelka_noise_2024} and \citet{muhojoki_benchmarking_2024} demonstrated that for modern mobile laser scanners and typical tree detection distances (\qtyrange[range-phrase = --]{1}{20}{\metre}) the relation is mostly linear. \citet{muhojoki_benchmarking_2024} and \citet{faitli_integration_2024} calibrated the bias by measuring the bias as a function of measurement distance in one sample plot and fitting a linear model to the data. The (system-specific) models were subsequently used for correcting the measurements from other sample plots, some of which had a drastically different tree species composition. Although the method proved effective---eliminating \qtyrange[range-phrase = --]{50}{99}{\percent} of the bias---the large variation demonstrates that the method can be improved. For example, it does not take the scan angle, tree species (reflectance, bark roughness), potentially heteroscedastic stochastic measurement noise or tree diameter into account.

The effect of noise and tree diameter on the accuracy of MLS/TLS-based tree diameter measurements  has been studied in the past literature, again with conflicting findings. \citet{gollob_forest_2020, stovall_comparing_2023} and \citet{kuzelka_noise_2024} reported a decreasing bias; a large positive bias for small, under 10 cm diameter trees and a small negative bias for large trees, whereas \citet{perez-martin_assessment_2021} found the bias to increase as the tree diameter increased. To correct the bias related to DBH, \citet{gollob_forest_2020} developed a model that dependent on tree diameter, mean intensity, and standard deviation and residual error whose coefficients were calibrated with manual DBH measurements. In addition to models that correct for beam width- or diameter-induced bias, \citet{verhoeven_fitting_2025} developed an uncertainty-based circle fitting algorithm for point clouds with heteroscedastic and potentially scan angle dependent noise. Although the bias compensation model can benefit even the state-of-the-art laser scanning systems \citep{muhojoki_benchmarking_2024}, it is even more important for low-cost ones \citep{mokros_novel_2021, cakir_3d_2021, gollob_measurement_2021, hu_development_2021, mcglade_potential_2022, balestra_advancing_2024, howie_measuring_2024}, due to their wider beams and higher noise.

This study investigates the effect of beam divergence and measurement noise on the estimation of tree diameter from mobile laser scanning point clouds. We simulated tree diameter measurements by a time-of-flight laser scanner to demonstrate the measurement distance and tree diameter dependency of the stem curve measurement bias and to find the source of conflicting bias observations in the previous literature. The findings were validated with both an artificial target cylinder and trees from sample plots using terrestrial laser scanning as a reference. Lastly, we developed a novel simulation-based bias compensation method that requires no prior calibration and corrects the diameter estimation bias based both on the measurement distance and the target tree diameter.

\section{Material and Methods}
\label{sec: materials and methods}
This study is broken into three parts:
\begin{itemize}
    \item Simulation study
    \item Tests on an artificial cylinder under laboratory conditions
    \item Validation tests on forest sample plots
\end{itemize}

In this section, simulations experiments are first described in Section~\ref{sec: simulations}, followed by a description of the used laser scanners, the forest sample plots, and the artificial cylinder in Section~\ref{sec: test sites and laser scanners}. Finally, the process of acquiring the stem curve and other tree stem attributes from the point clouds is presented in Section~\ref{sec: point cloud processing}, and the implementation of bias compensation in Section~\ref{sec: implementation}.

\subsection{Simulations}
\label{sec: simulations}
\begin{figure}
	\centering
	\begin{subfigure}[t!]{0.48\textwidth}
        \caption{}
        \includegraphics[width = \textwidth]{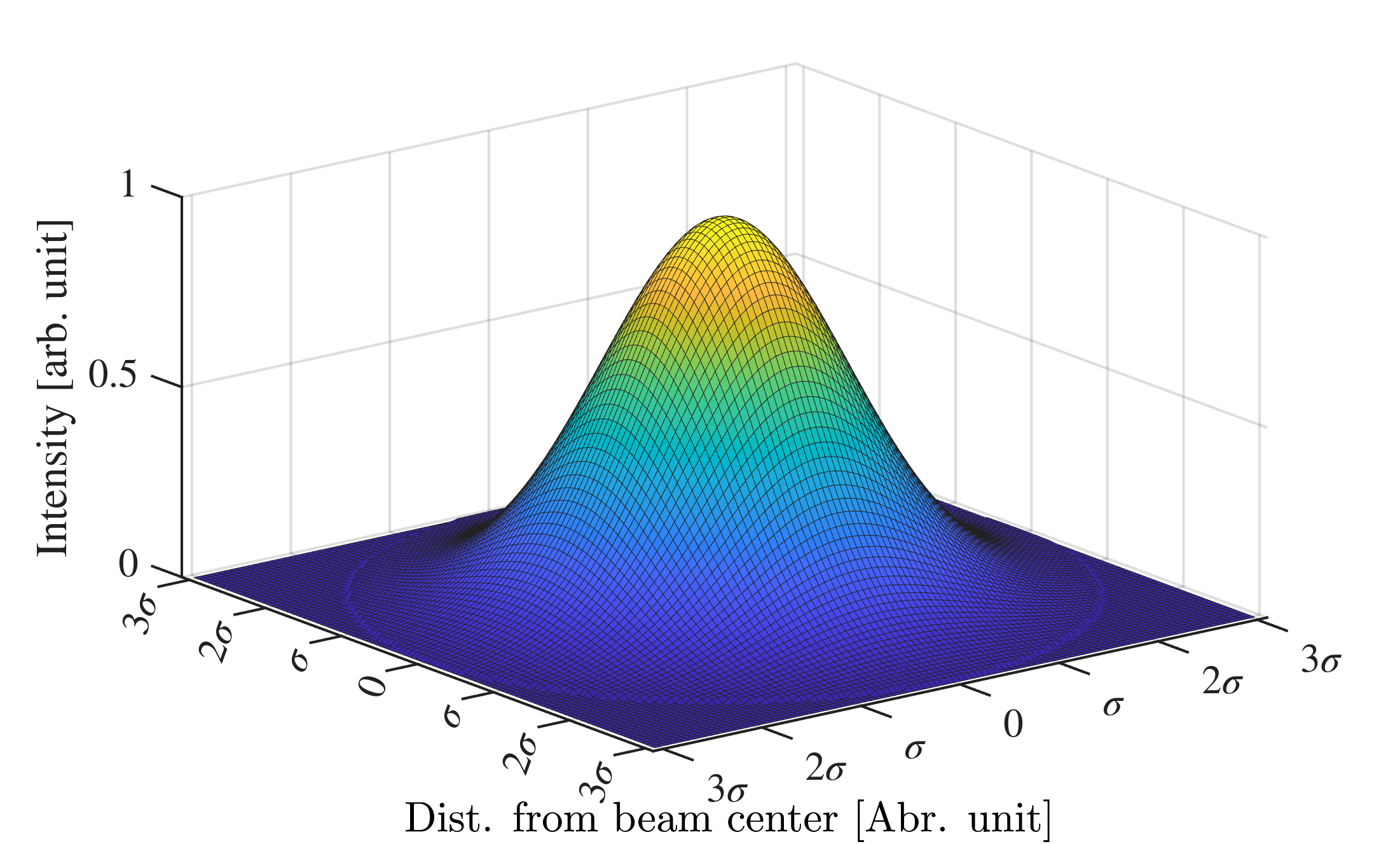}
    \end{subfigure}
    \hfill
    \begin{subfigure}[t!]{0.48\textwidth}
        \caption{}
        \includegraphics[width = \textwidth]{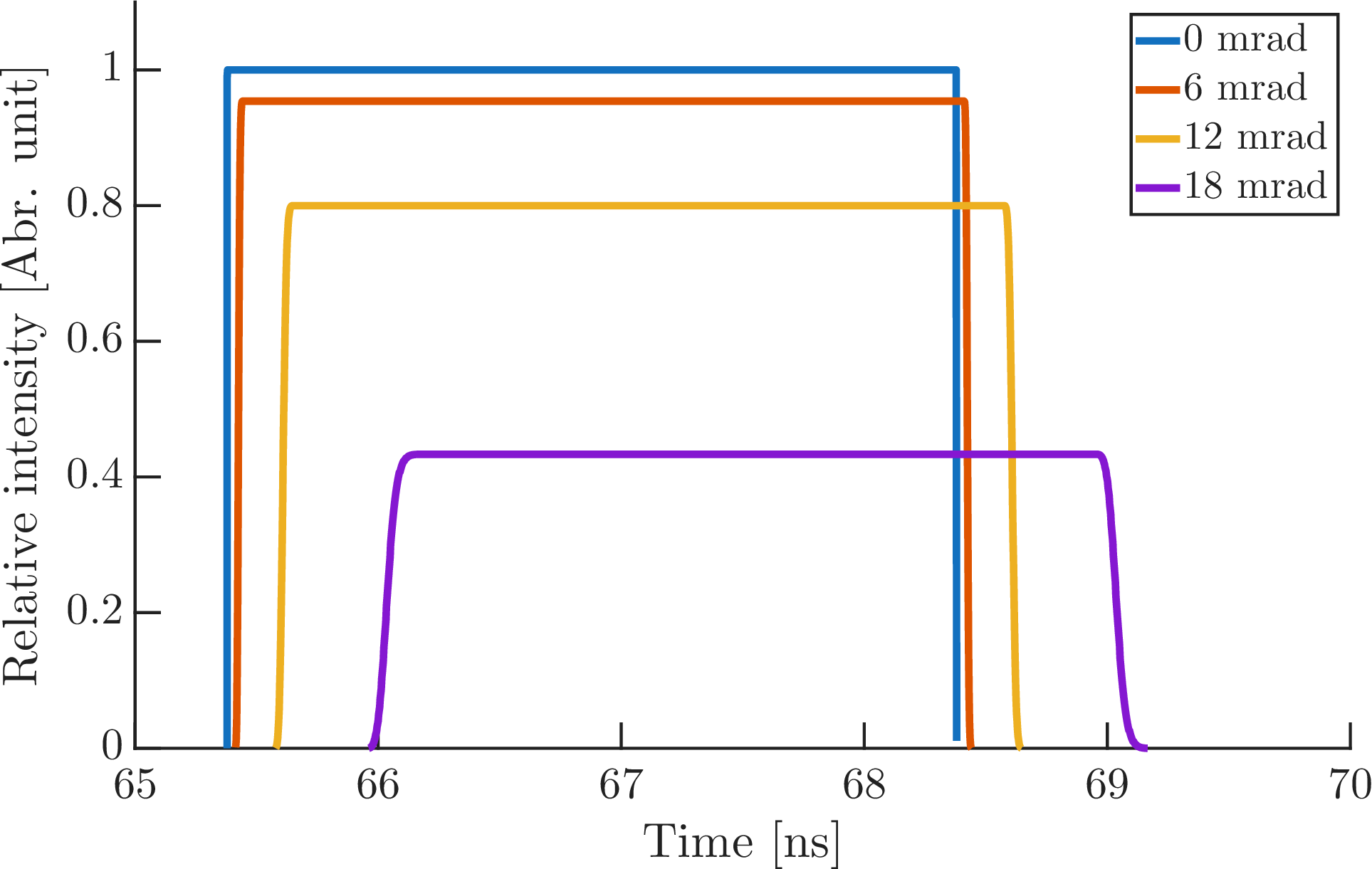} 
    \end{subfigure}
    \begin{subfigure}[t!]{0.48\textwidth}
        \caption{}
        \includegraphics[width = \textwidth]{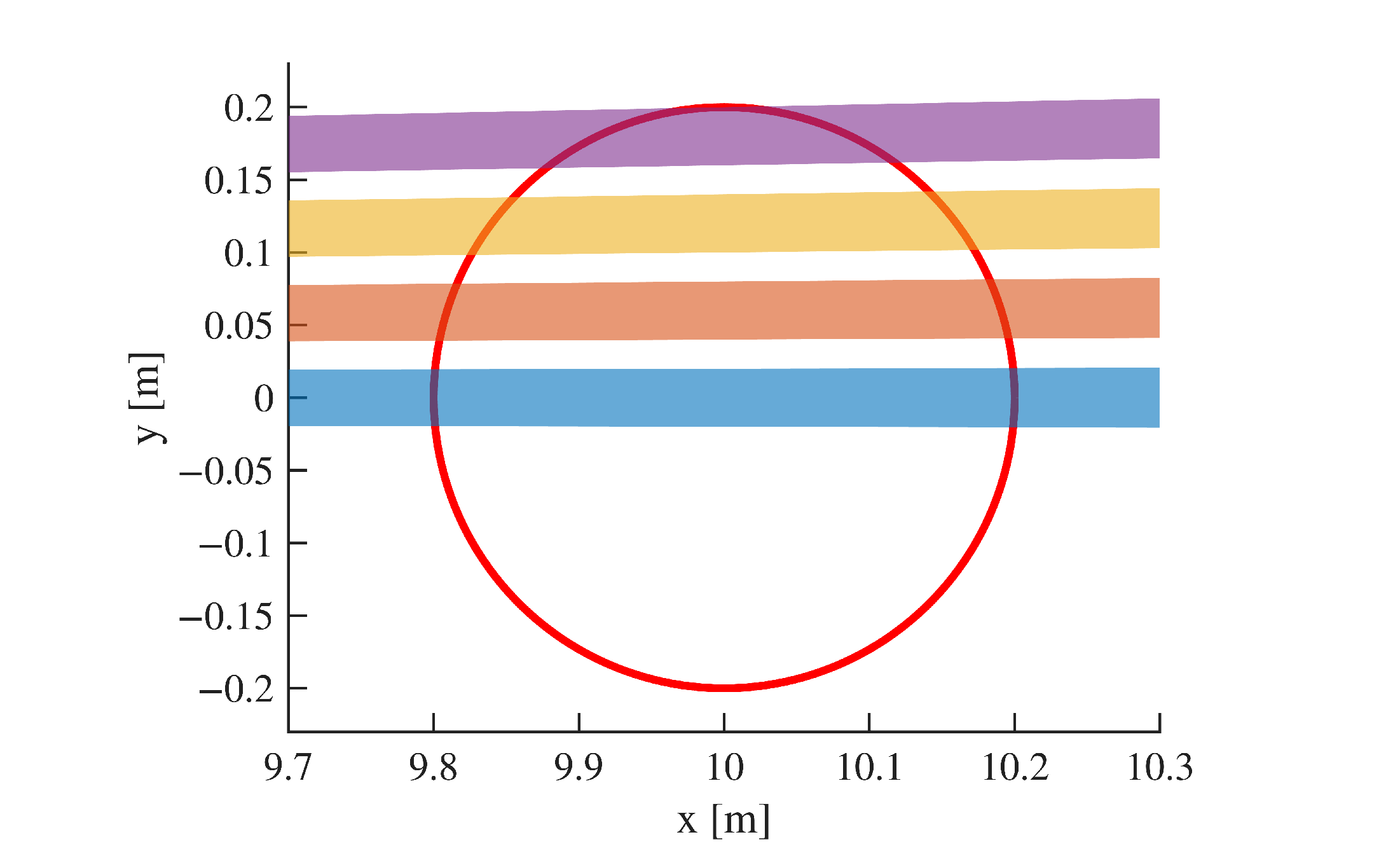}
    \end{subfigure}
   \caption{ \label{fig:  simulation illustration} a) Beam intensity distribution exiting the sensor. b) Return signal for measurements illustrated in c) (top view, target cylinder in red). The edges are smeared by the finite beam hitting a curved surface. The laser pulse was fired from the origin at an angle of 0, 6, 12 and 18 mrad w.r.t. $x$-axis. The pulse was 3 ns long with divergence of 2 mrad.}
\end{figure}

The simulations followed the approach presented by \citet{forsman_bias_2018}. The beam was assumed to be Gaussian and circular, and the beam waist was ignored. At the scanning location, the beam width was \textit{beam width at exit}, $W_0$, at $2\sigma$, as specified by the laser scanning system manufacturer, and the beam was simulated up to $3\sigma$. As $W_0$ was not given for the MLS system used in this study, we used a representative value for a modern lidar of $W_0 = \qty{5}{mm}$. The $2\sigma$ definition of the beam width, $W$, is used throughout this paper. As the targets are much further away from the laser scanner than the Rayleigh length, and the divergence of the beam $\Theta$ is of the order of \qtyrange[range-phrase = --]{1}{10}{mrad}, the beam width at distance of $d$ from the scanner is simply given by $W(d) = W_0 + d\Theta$, where we have employed the small-angle approximation $\mathrm{tan}(\Theta) \approx \Theta$. The beam cross-section was spatially discretized into a 151-by-151 grid. The intersection point and angle between the target cylinder and each sub-beam were calculated, and the reflected pulse was assumed to follow Lambert's cosine law $I(\theta) = I_0 \mathrm{cos}(\theta)$, where $\theta$ is the incident angle. Reflectivity and absorption were ignored. The time of incidence was determined from the intersection point and the return signal was calculated by temporally discretizing the beam between the first incidence to the last beam leaving the surface into 1000 segments and aggregating the returned intensity. The location of the measured point was determined to be in the center of the beam, $d = c\Delta t/2$ distance away from the origin of the beam, where $c$ is the speed of light and $\Delta t$ the time difference from the beam leaving the sensor to when the set proportion of the amplitude of the pulse was returned to the scanner. This proportion threshold is called \textit{trigger threshold} ($t_\mathrm{thresh}$), and is calculated with respect to the amplitude of the pulse exiting the lidar. The details of $t_\mathrm{thresh}$ are not typically given by the sensor manufacturers, and therefore we used a default value of $t_\mathrm{thresh} = 0.1$, as suggested by \citep{forsman_bias_2018}. We assumed that $t_\mathrm{thresh}$ does not depend on the measurement distance, as the ranging accuracies stated by the manufacturer would not be possible without some sort of range correction due to range-walk \citep{yang_double_2019}. Furthermore, adjusting $t_\mathrm{thresh}$ can be used to simulate different levels of target reflectivity, since increasing $t_\mathrm{thresh}$ w.r.t.\@ the ideal reflection is roughly equivalent to decreasing the reflectivity of the target. The simulation of the gaussian beam and the effect of the finite beam width on the return pulse are illustrated in Fig.~\ref{fig:  simulation illustration}.

\begin{figure}
	\centering
	\begin{subfigure}[t!]{0.48\textwidth}
        \caption{Homoscedastic noise}
        \includegraphics[width = \textwidth]{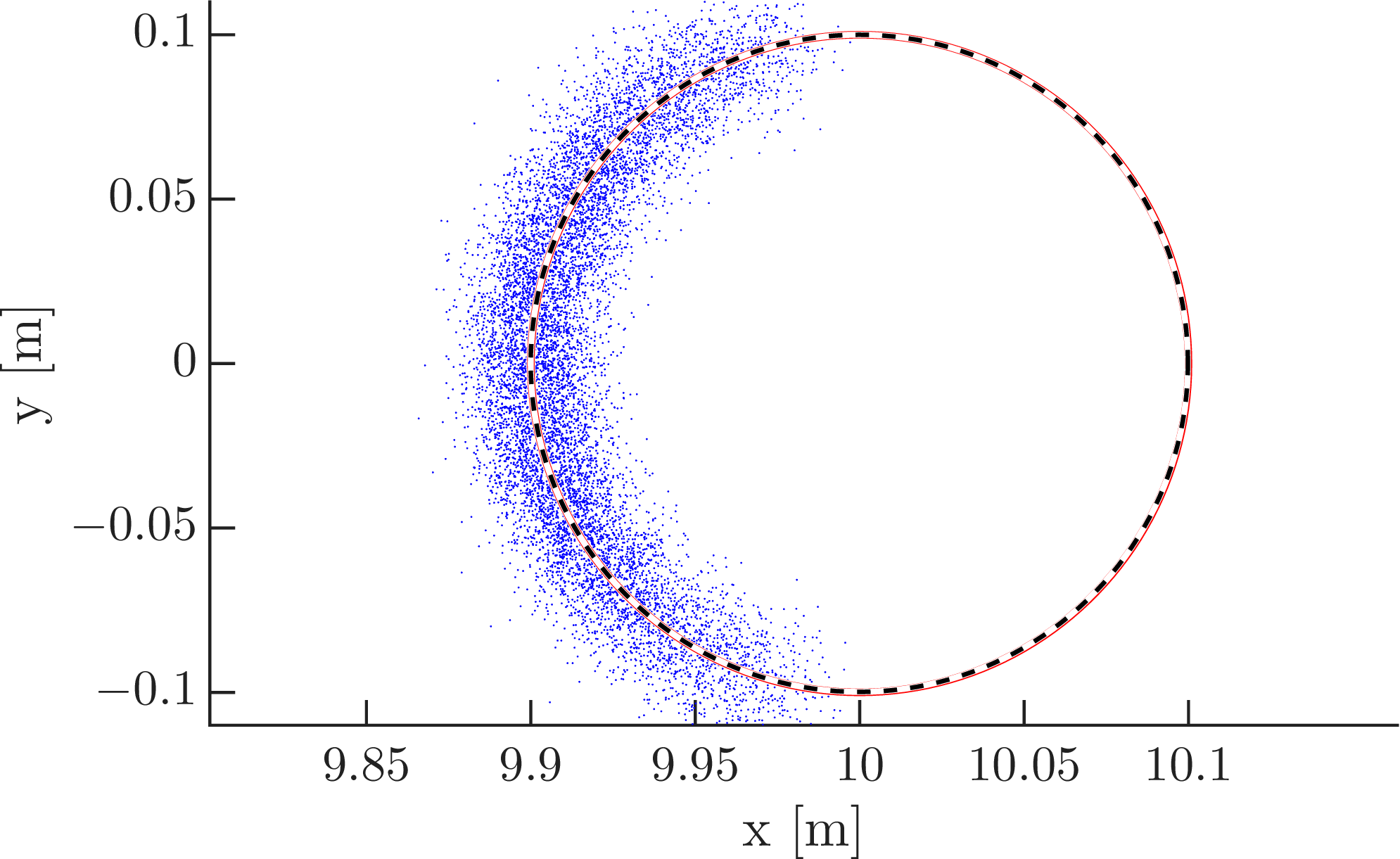}
    \end{subfigure}
    \begin{subfigure}[t!]{0.48\textwidth}
        \caption{Heteroscedastic noise}
        \includegraphics[width = \textwidth]{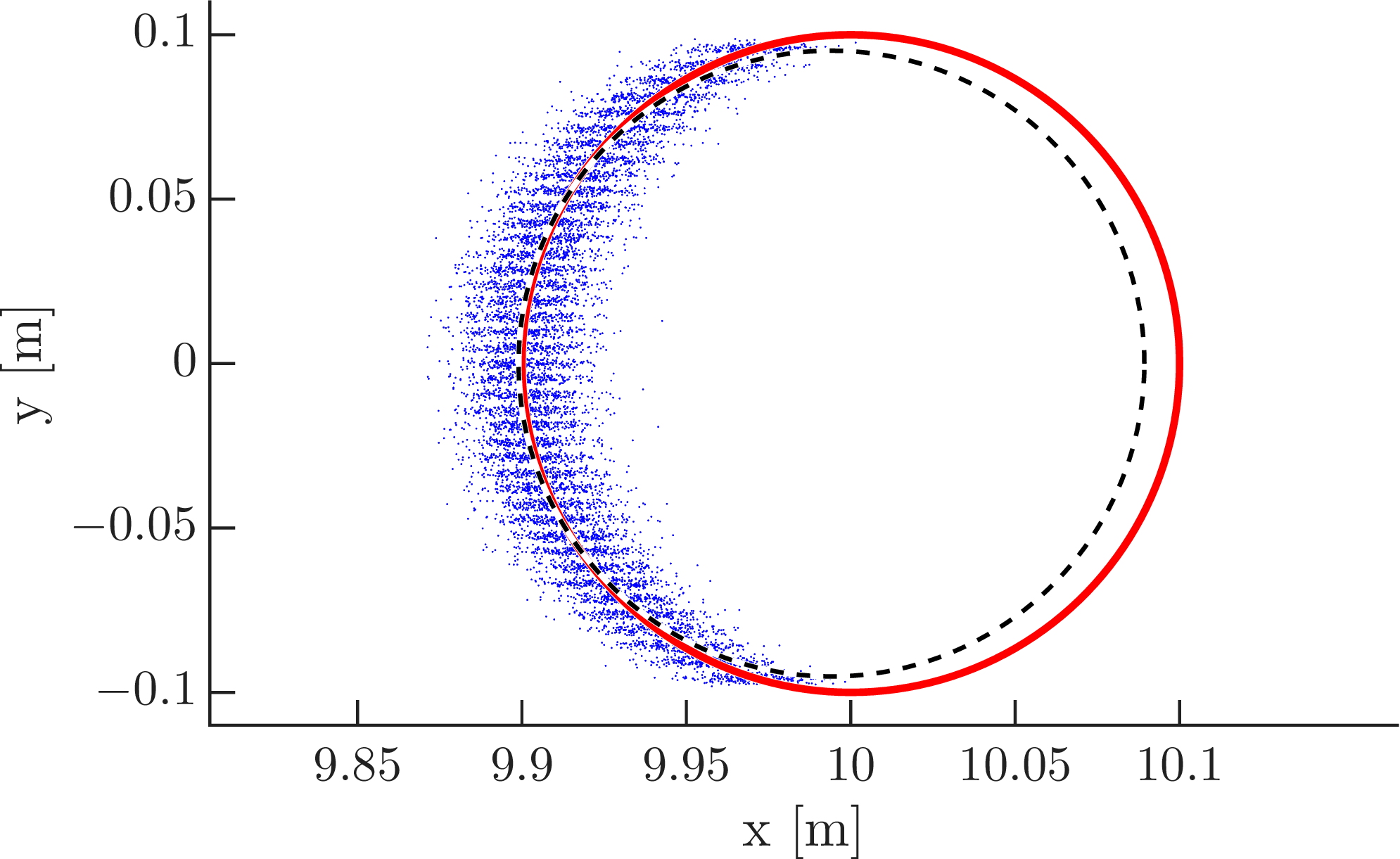} 
    \end{subfigure}
   \caption{ \label{fig:  noise bias illustration} Illustration of increasing circle fit bias as the heteroscedasticity of the noise increases. a) Homoscedastic noise with $\sigma_x= \sigma_y = 1$ cm. b) Heteroscedastic noise with range noise with $\sigma_r=1$ cm and angle noise of $\sigma_\theta=0.1$ mrad. The blue dots are the simulated laser data points, the red circle represents the cylinder and black dashed line the fitted circle. The lidar is located 10 meters left from the center of the tree.}
\end{figure}

In addition to the beam divergence, we simulated the uncertainty in the range and angle measurements. Although computationally efficient algebraic circle fit can be performed without inducing bias if the noise is homoscedastic \citep{al-sharadqah_error_2009}, i.e.,\@ the noise has the same variance in all directions, in general the measurement noise of a lidar is heteroscedastic \citep{verhoeven_fitting_2025}, which can cause bias in the circle fit, as demonstrated in Fig.~\ref{fig:  noise bias illustration}. The noise was simulated by adding normally distributed noise with $\sigma_r$ to the range measurement and $\sigma_\phi$ to the azimuthal angle (origin at the lidar, the laser beams rotate about the $z$-axis).

We utilized the simulations to investigate both isolated and combined effects of finite beam width and ranging precision on the bias of estimating the diameter of the cylinder (tree stem). We compared the findings to lidar measurements on both a smooth artificial cylinder and (forest) sample plots. Furthermore, the simulations form the basis for the bias compensation method described in Section~\ref{sec: implementation}. 

\subsection{Test sites and laser scanners}
\label{sec: test sites and laser scanners}

The validation experiments were performed in Evo, Finland ($\ang{61.19}$ N, $\ang{25.11}$ E) on a total of 9 $\qty{32}{\metre} \times \qty{32}{\metre}$ sample plots representing varying boreal forest structures. The sample plots were categorized into \textit{easy} and \textit{difficult}, based on the the stem density and tree species composition. The sample plots categorized as easy were sparse ($<1000$ stems/ha), and contained mainly Scots pine (\textit{Pinus sylvestris} L.) that typically features few lower-hanging branches and a highly circular and straight stem. Therefore, most sources of error other than the laser scanners themselves were expected to be eliminated. In contrast, the difficult sites consisted of mixed forests dominated by Norway spruce (\textit{Picea abies} (L.) H.Karst), silver and downy birches (\textit{Betula pendula} Roth, \textit{Betula pubescens}, Erth.) and Aspen (\textit{Populus Tremula} L.) with mixtures of Scots pine and a few other deciduous trees. The density of trees at the difficult sites ranged from approximately 1000 to 2500 stems/ha. In these sites, occlusion of stems, curved or otherwise deformed trees, and the ambiguity of matching point cloud-derived trees with the reference stem map due to close proximity of trees play a major role in the error analysis.

\begin{figure}[htbp]
    \begin{subfigure}{0.49\textwidth}
        \centering
        \caption{MLS}
        \includegraphics[width=\linewidth]{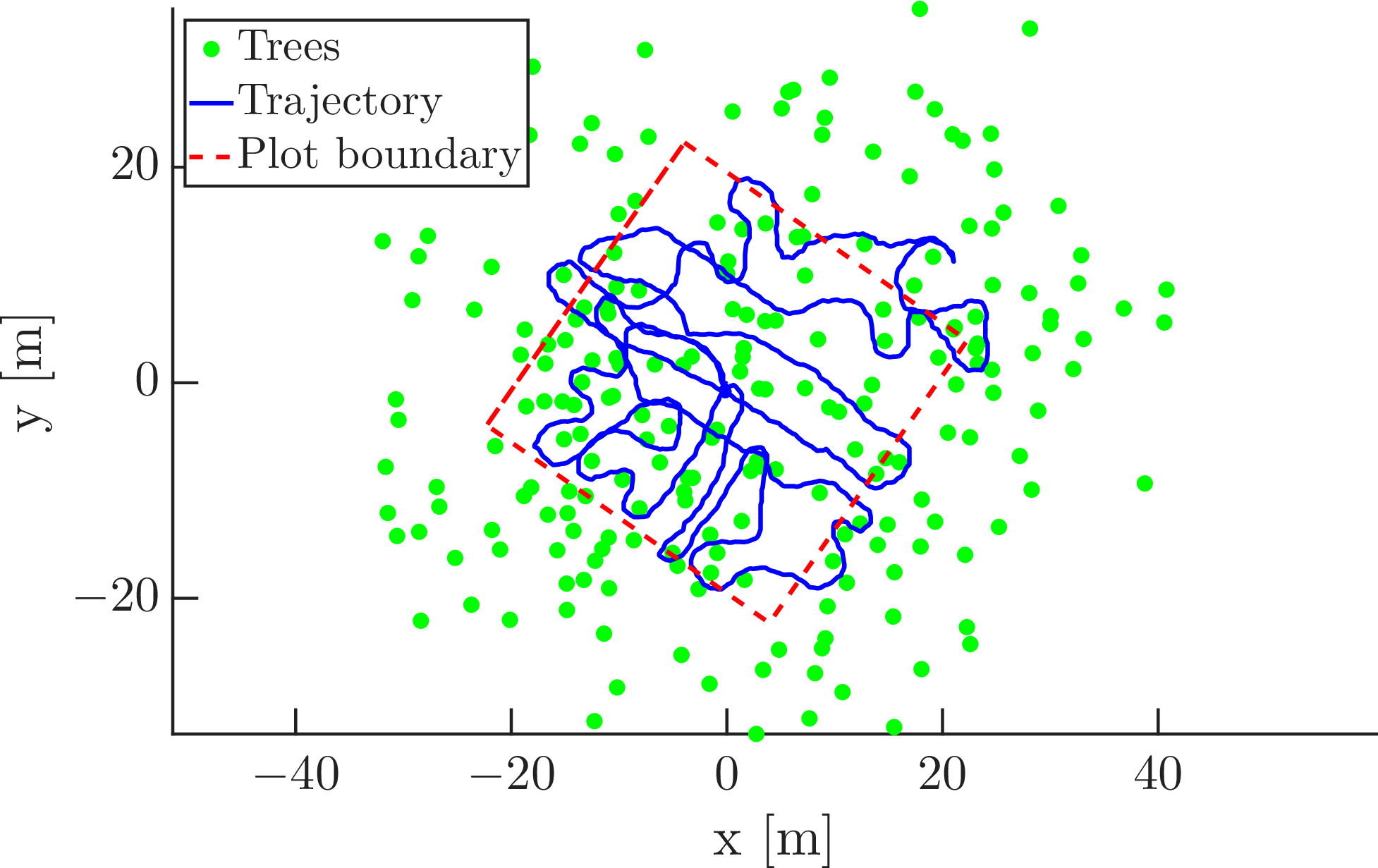}
    \end{subfigure}
    \hfill
    \begin{subfigure}{0.49\textwidth}
        \centering
        \caption{TLS}
        \includegraphics[width=\linewidth]{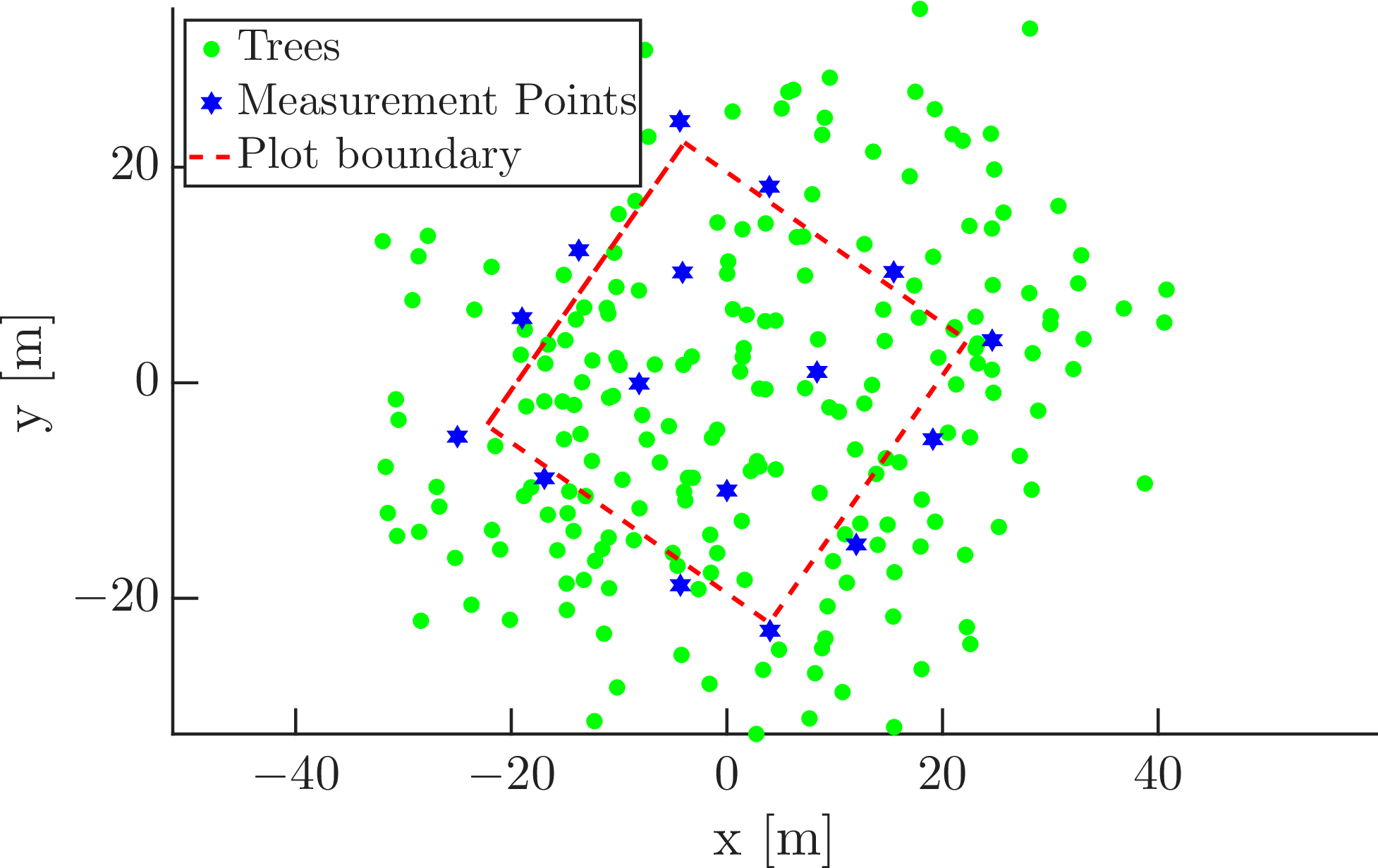}
    \end{subfigure}
    \caption{(a) Example MLS measurement trajectory. (b) Example 16-point TLS measurement setup. The plot boundaries are presented as red dotted line.}
\label{fig: example measurement setups}
\end{figure}

The nine sample plots were measured with a Faro Orbis (FARO Technologies Inc., Lake Mary, Florida, USA) mobile laser scanner on 22\textsuperscript{nd} of May 2025. The data were processed into point clouds using Faro Connect software (version 3 2024) with the "forest" setting enabled for simultaneous localization and mapping (SLAM). For reference stem curves, the same plots were measured a Riegl VZ-400i (Riegl GmbH, Horn, Austria) TLS system from 16 scanning positions per plot on 6\textsuperscript{th}--20\textsuperscript{th} of May 2025. The reference measurements were registered and processed into point clouds using RiSCAN PRO software (version
2.20.1). The scanning positions were placed in a regular $4\times4$ configuration that was slightly adjusted, if necessary, due to local topography or tree positions. At every other measurement point, an additional scan with the scanner tilted $\ang{90}$ was taken for a hemispherical scan for a more complete vertical field of view. An example TLS setup and MLS measurement trajectory are presented in Fig.~\ref{fig: example measurement setups}. Selected key laser scanner specifications are presented in Table~\ref{tab: scanner stats}

\begin{table}[ht]
\centering
\begin{threeparttable}
\caption{Lidar specifications}\label{tab: scanner stats}%
\begin{tabular}{@{}lccccc@{}}
\toprule
Scanner/property & Faro Orbis & Riegl VZ-400i \\
\midrule
Divergence [mrad] & $0.7\times 1.7$\tnote{a} & 0.35 \\
$1\sigma$ Ranging precision [mm] & 5 & 3 \\
Ranging accuracy [mm] & $\pm10$ & 3--5\tnote{b}\\
PRR\tnote{c} [kHz] & 20 & 1200 \\
Angular Resolution [mrad] & 6.3 & 0.7 \\
Wavelength [nm] & 905 & 1550 \\
\bottomrule
\end{tabular}
\begin{tablenotes}
\item[a] Vertical $\times$ Horizontal.
\item[b] 3D accuracy @ \qtyrange[range-phrase = --]{50}{100}{\metre}
\item[c] (Used) Pulse Repetition Rate.
\end{tablenotes}
\end{threeparttable}
\end{table}

\subsection{Cylinder measurements}
\begin{figure}[ht]
	\centering	
    \includegraphics[width = 0.5\textwidth]{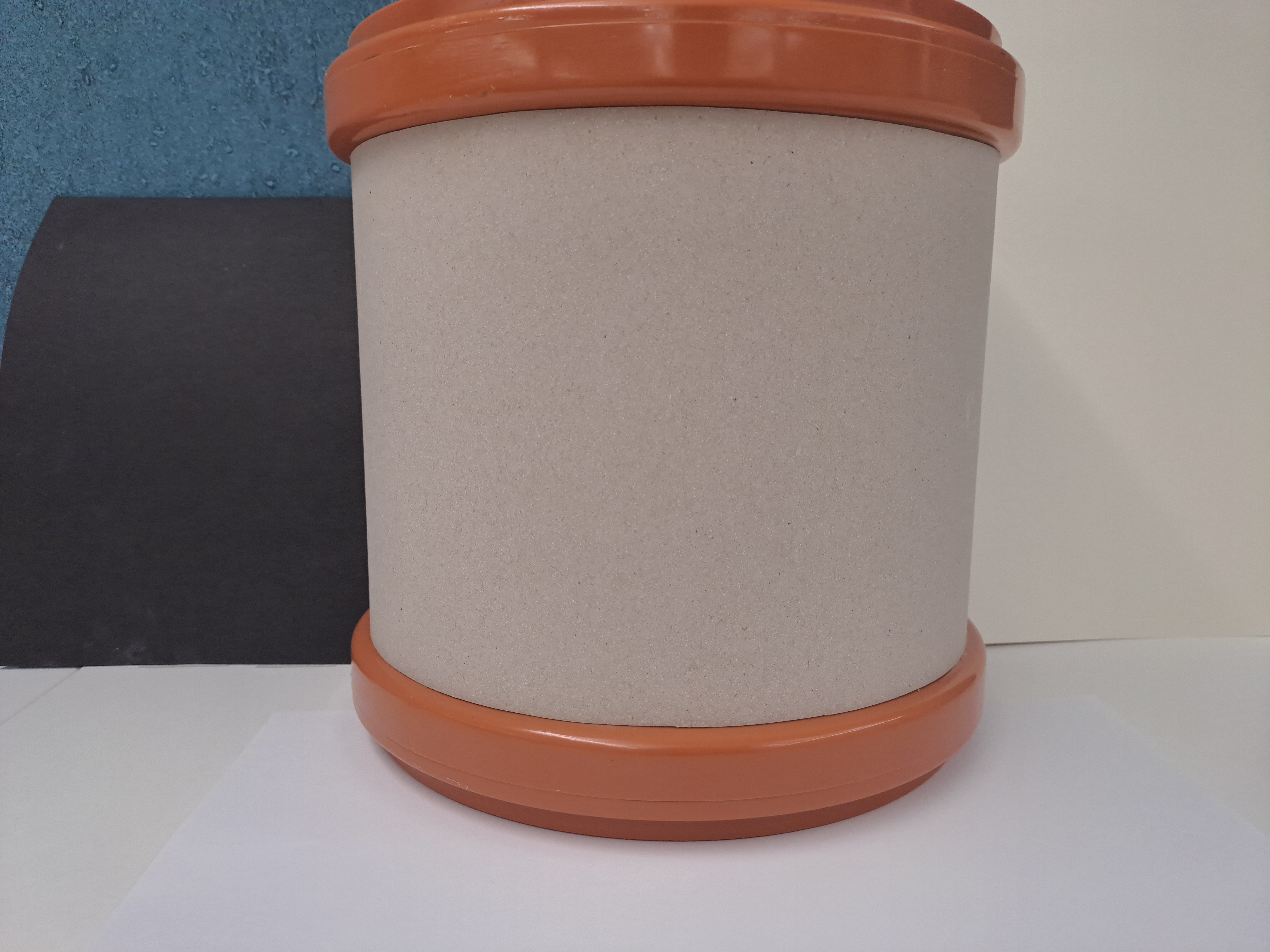}

   \caption{ \label{fig: target illustration} 
    The artificial target. The brown cardboard is attached to the target and the black and white cardboard are on the left and right of the target, respectively.
   }
\end{figure}

To validate the simulations in a controlled environment, we measured a cylinder with varying reflectance at various ranges. A cylinder approximates the shape of a tree stem, while its geometric properties are precisely known, making it suited for use in both method development and validation. The cylinder had a uniform diameter of \qty{210}{mm} and height of \qty{155}{mm} and was covered with thin colored cardboard to simulate different tree species, as illustrated in Fig.\ref{fig: target illustration}. The colors of the paper were brown, white, and black, where brown and white represent the range of reflectivities found in the trees of a boreal forest, and black was selected as a low reflectance comparison. Data were collected at a distance of \qtyrange[range-phrase = --]{1}{15}{\metre} from the cylinder in increments of \qty{1}{\metre}. The Faro Orbis scanner was held in the hand of the operator at the same height as the cylinder, and each distance was measured by holding the scanner still for approximately \qty{60}{\second} and then stepping back to the next measurement point. The uniform middle part of the cylinder was manually segmented from the point clouds and the cylinder diameter was calculated with a simplified version of the full stem curve software (see Section~\ref{sec: point cloud processing}): The point clouds were divided into 3-second segments and each segment was first filtered with RANSAC (random sample consensus \citep{fischler_random_1981}) circle fitting, followed by algebraic "hyperaccurate" circle fitting \citep{al-sharadqah_error_2009} to obtain the diameter estimates. The diameters were binned into distance bins based on the trajectory time stamps during which the scanner was stationary. For each measurement distance, the distance to the cylinder was defined as the distance between the average scanner position and the center point of the cylinder, and the median of the fitted circle diameters was used as the measured cylinder diameter.

\subsection{Point cloud processing and reference}
\label{sec: point cloud processing}
\begin{figure*}[ht]
	\centering	
    \includegraphics[width = 0.98\textwidth]{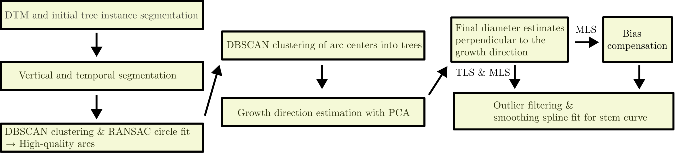}

   \caption{ \label{fig: process illustration} 
    Process flow for the stem curve and DBH estimation. In the final step, MLS data is processed both with and without the bias compensation.
   }
\end{figure*}

Both the MLS and reference TLS data collected from sample plots were initially processed into stem curves using the method presented by \citet{hyyppa_under-canopy_2020}. The method segments the point cloud into temporal and spatial segments and extracts \textit{arcs}, i.e.,\@ consecutive laser hits to the tree stem forming an arc-shaped cluster of points, using DBSCAN \citep{ester_density-based_1996}. The arcs were filtered using RANSAC circle fitting, and preliminary diameter estimates were acquired with an algebraic "hyperaccurate" circle fitting method \citep{al-sharadqah_error_2009}, after projecting points to a plane perpendicular to the primary growth direction of the tree. Finally, a smoothing spline and an outlier filter were applied to obtain the diameter estimates at different heights (stem curve). For a detailed description of the method, the reader is directed to \citet{hyyppa_under-canopy_2020}. The parameters used in this study are described in \citet{muhojoki_benchmarking_2024}. Most importantly, we used a temporal segment duration of \qty{3}{\second}, which allows extracting high-quality arcs and determining the arc-scanner distance for later distance-based corrections. During each 3-second time interval, the trajectory provided by the SLAM-corrected IMU is stable, allowing the extraction of high-quality arcs. However, the trajectory tends to drift over longer time periods without external reference, such as GNSS. Furthermore, in a typical MLS measurement setting in a forested environment, the scanner only moves a few meters during the \qty{3}{\second} period, and thus the average $x$- and $y$-coordinate is a reasonable approximation for the scanner location. The following section details a simulation-based bias compensation method used to further process the MLS arcs. Uncompensated MLS data was used as a comparison and for investigating the distance and tree diameter dependency of the measurement bias. The process, including the optional bias compensation, is illustrated in Fig.~\ref{fig: process illustration}.

\subsection{Implementation of the bias compensation}
\label{sec: implementation}

We developed a simulation-based bias compensation method to demonstrate the feasibility of calibration-free bias compensation, relying only on laser scanner specifications. As running a simulation for each arc separately was too slow with the current simulation implementation, we tabulated simulated measurements of trees with a diameter from 75 mm to 1200 mm at intervals of 25 mm, for distances ranging from 1 m to 20 m at 1 m intervals. We applied heteroscedastic, normally distributed noise to the measurement and recorded the mean measured cylinder diameter from 10000 samples and the measurement distance between the scanner and the center of the cylinder. The measured cylinder radius was calculated using the same RANSAC circle fit followed by the "hyperaccurate" fit as described in Section~\ref{sec: point cloud processing}, and the average measured cylinder radius is thereafter referred to as \textit{simulated radius}.

The important parameters for the simulation are trigger threshold $t_\text{thresh}$, beam divergence $\Theta$ and standard deviations of range and azimuthal angle noise $\sigma_r$ and $\sigma_\phi$. The range precision is typically provided by the manufacturer, but may vary depending on the measurement conditions and target properties. Beam divergence is also often given. However, azimuthal precision, $t_\text{thresh}$ and further details concerning range calculations from the signal are not. Therefore, it is crucial that the method is stable, and can correct bias in diameter measurements even if the simulation values are slightly wrong. For baseline values, we selected $t_\text{thresh} = 0.1$ based on the arguments presented in \citet{forsman_bias_2018}. The Faro Orbis has a reported nominal range accuracy of $\pm 10$ mm with a precision of $\sigma_r = 5$ mm ($1\sigma$) and a nominal azimuthal resolution of \qtyrange[range-phrase = --]{2}{7}{mrad}. As the noise should be lower than the resolution, we chose $\sigma_\phi = 0.5$ mrad. The Faro Orbis has a nominal beam divergence of $\qty{0.7}{mrad} \times \qty{1.7}{mrad}$ \citep{kokeza_comparative_2026}. As the lidar of the Faro Orbis is mounted on a rotating platform that rotates approximately at \qty{1}{\hertz}, the arcs can contain points from up to 3 rotations of the lidar, i.e.,\@ the elliptical footprints will be oriented differently within an arc, the beam was modeled as a cone, with baseline divergence of the $\Theta=\qty{1.1}{mrad}$, close to the geometric average divergence of $\sqrt{0.7 \ \mathrm{mrad} \times 1.7 \ \mathrm{mrad}}= 1.09$ mrad.

To correct the bias of an individual arc, we first calculated the distance between the scan location and the arc, where the location of the arc was defined as the center of the initial circle fit, and the scan location as the average scanner trajectory coordinate between the detection of the first and the last point in the arc. As the points are separated by at most \qty{3}{\second}, the trajectory point are typically separated by at most few meters due to slow walking speed in the forest. Subsequently, the two closest simulated mean diameters were selected from two closest simulated measurement distances, and the new diameter estimate was computed as the weighted average based on the difference between the measured and simulated distance and diameter. The weights $w_{d_1}$ and $w_{d_2}$ for nearest distances $d_1$ and $d_2$ were calculated as: 

\begin{equation} \label{eq: distance weights}
    w_{d_n} = \frac{|d_m - d|}{|d_m - d_n|},   
\end{equation}

where $d$ is the measurement distance and $n,m\in\{1,2\}, n\neq m$. Similarly, for both distances, the two nearest simulated radii $r_{n,1}$ and $r_{n,2}$ to the measured radii $r$ were determined and their weights were calculated as:

\begin{equation} \label{eq: radius weights}
    w_{r_{n, i}} = \frac{|r_{n, j} - r|}{|r_{n, j} - r_{n, i}|}, 
\end{equation}

where $i\in\{1,2\}$. Subsequently, the Eqs.~\eqref{eq: distance weights} and \eqref{eq: radius weights} were utilized in calculation of the corrected radius $r_{\mathrm{corr}}$:

\begin{equation} \label{eq: final correction}
    r_{\mathrm{corr}} = w_{d_1} \left( w_{r_{1,1} } \hat{r}_{1,1} + w_{r_{1,2} } \hat{r}_{1,2} \right) + w_{d_2} \left( w_{r_{2,1} } \hat{r}_{2,1} + w_{r_{2,2} } \hat{r}_{2,2} \right),
\end{equation}

where $\hat{r}_{n,i}$ represents the cylinder radius corresponding to the simulated radius $r_{n,i}$. After all arcs of a single tree have been bias-corrected using Eq.~\eqref{eq: final correction}, the smoothing spline and outlier filter are re-applied, as described in Section~\ref{sec: point cloud processing} and illustrated in Fig.~\ref{fig: process illustration}.

\section{Error Analysis and Evaluation of the the Bias Compensation}
\label{sec: evaluation metric}

The performance of the bias compensation method in the sample plots is compared to the uncompensated results using root-mean-square error (RMSE) and bias, defined as:

\begin{align}
    \text{RMSE} = &\sqrt{ \frac{1}{N} \Sigma_{i = 1}^{N} (D_{i} - D_{\mathrm{ref},i}})^2 , \label{eq: RMSE}\\
    \text{Bias} = &\frac{1}{N} \Sigma_{i = 1}^{N} (D_{i} - D_{\mathrm{ref},i}), \label{eq: bias}, \\
\end{align}

 where, $D_i$ is the measured tree diameter from MLS and $D_{\mathrm{ref},i}$ is the corresponding reference value acquired with TLS. Bias was also used in error analysis with the artificial cylinder and simulation data, for which the reference value is known. For DBH, $D_i$ is the diameter at \qty{1.3}{\metre} above the ground for tree $i$. For the stem curve, the sum is calculated over all height intervals and trees. Ideally, with the used stem curve calculation parameters, each tree would have a diameter estimate from \qty{1.5}{\metre} to \qty{9.9}{\metre} above the ground in \qty{40}{\centi \metre} increments. However, due to occlusions, the diameter values could not always be reliably estimated throughout the whole height range. As discussed in Section~\ref{sec: point cloud processing}, the stem curve was smoothed with a spline, and therefore the stem curve was always defined between the highest and lowest height interval from which the diameter could be reliably estimated. However, we did not extrapolate the stem curve, and thus any non-overlapping region between the stem curve and its reference was ignored. 

 The diameter estimation error was studied as a function of stem diameter and measurement distance and also species- and sample plot-wise. In the sample plots, the effects of the tree diameter and measurement distance were isolated by binning each measured arc based on the average measurement distance and the reference tree diameter at the height of the arc. Only the arcs that were approximately parallel to the scanner at the height of \qtyrange[range-phrase = --]{1.5}{2.5}{\metre} were included to eliminate the potential effect of the vertical scan angle and bins with less than 50 arcs were discarded. For the isolated tree diameter experiment, the diameter bin centers were equidistant from \qty{10}{cm} to \qty{38}{cm} in \qty{4}{\centi\metre} increments. The results are presented for distance bins of 3, 5, 7, 10 and \qty{12}{\metre} ($\pm \qty{0.5}{\metre}$). Similarly, for studying the effect of the measurement distance on sample plots, the measurement distance bins were equidistant and 1-m-wide, with the bin centers ranging from \qty{1.5}{\metre} to \qty{24.5}{\metre}. The diameter bins were in this experiment were 5-cm-wide with the bin centers ranging from \qty{12.5}{\centi\metre} to \qty{27.5}{\centi\metre}. In addition, we employed a bin for all arcs smaller than \qty{10}{\centi\metre} and larger than \qty{30}{cm} in diameter.

\section{Results and Discussion}\label{sec: results}

\subsubsection{Simulation Experiments---Effect of Beam Divergence and Ranging Noise}

\begin{figure*}[htbp]
    \centering

    \begin{subfigure}{0.48\textwidth}
        \centering
        \caption{$\sigma_r = 0$, $D = 20$ cm}
        \includegraphics[width=\linewidth]{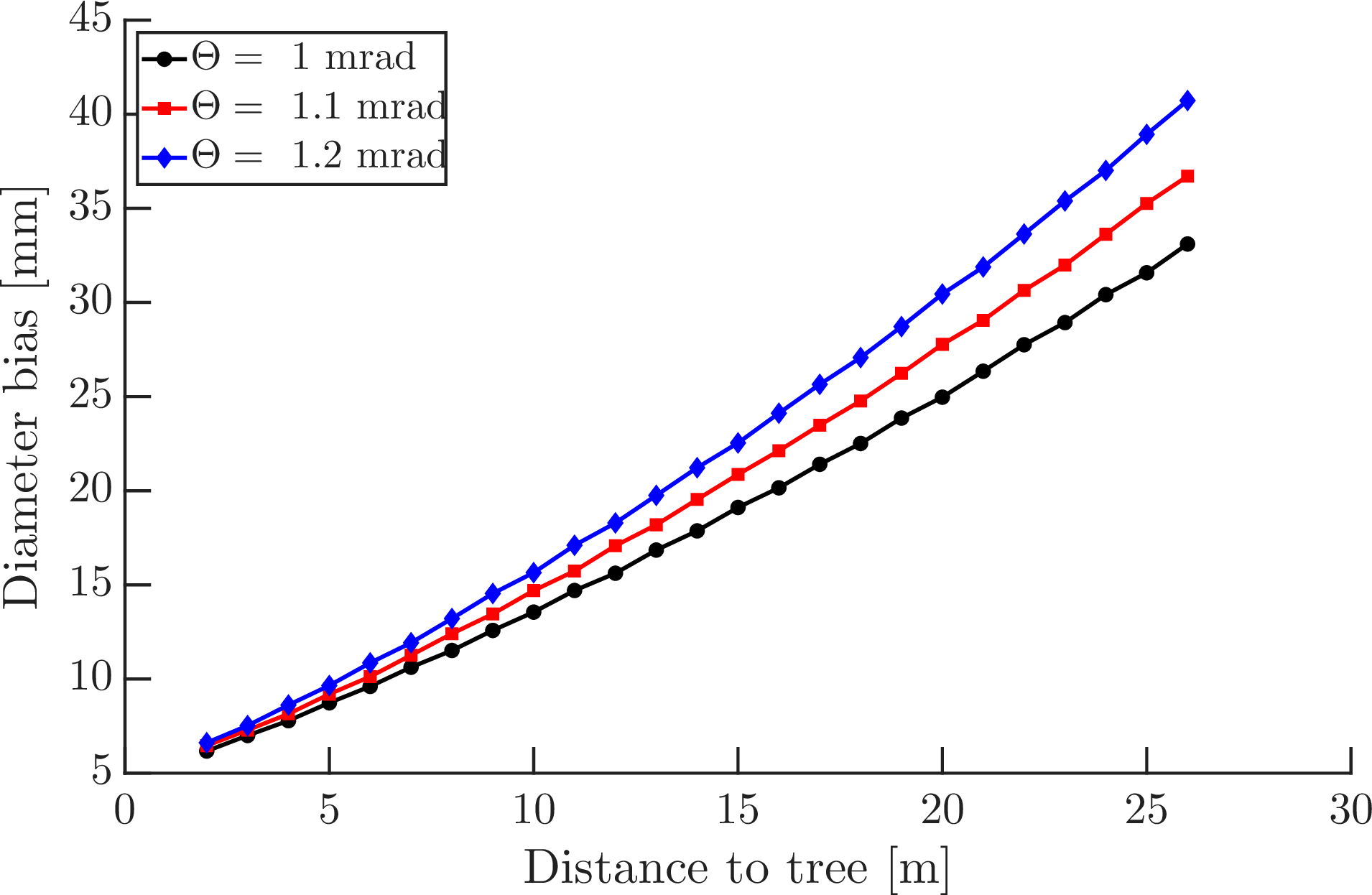}
    \end{subfigure}
    \hfill
    \begin{subfigure}{0.48\textwidth}
        \centering
        \caption{$\sigma_r = 0$, $d = 10$ m}
        \includegraphics[width=\linewidth]{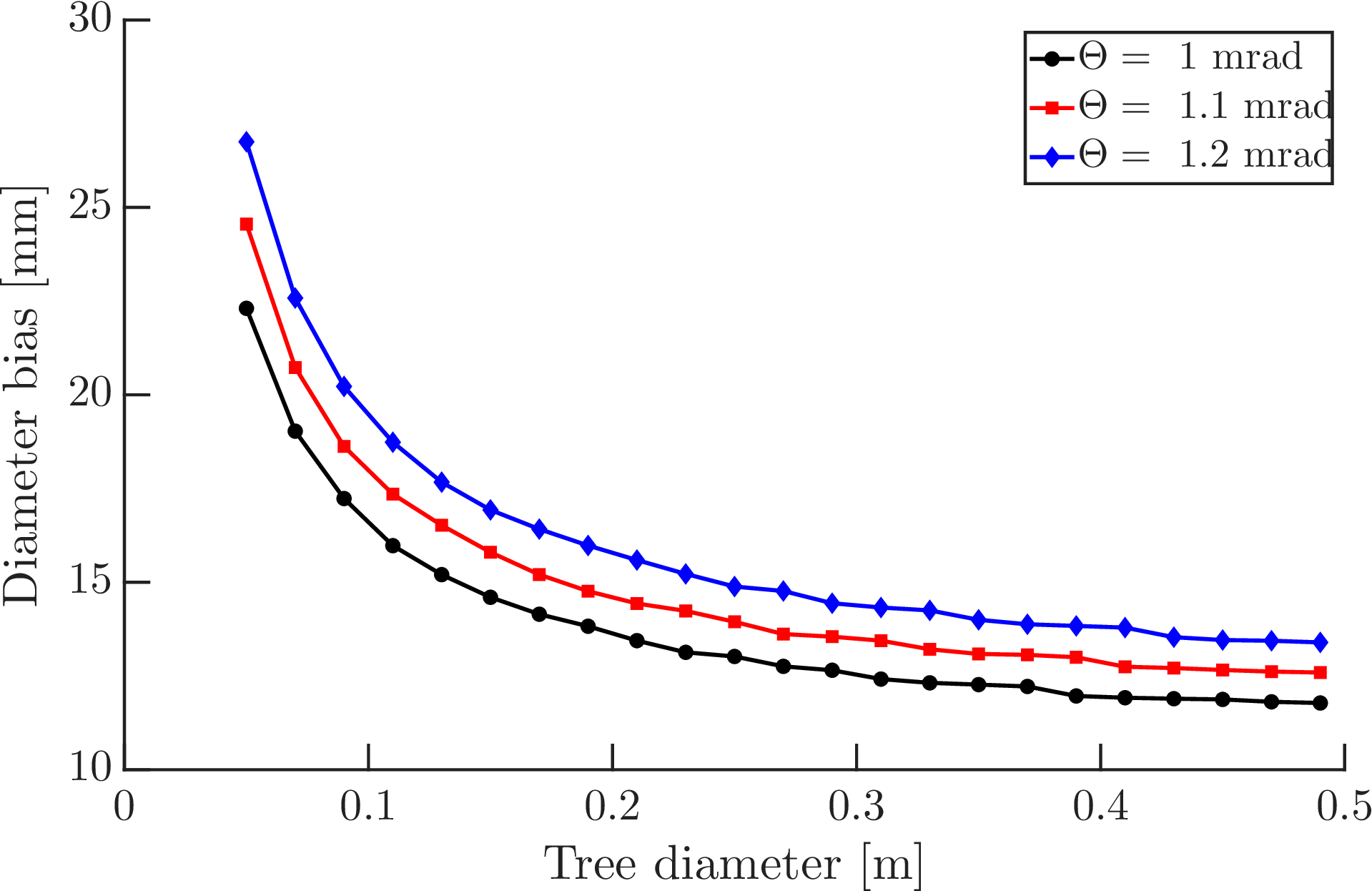}
    \end{subfigure}

    \vspace{0.5cm}

    \begin{subfigure}{0.48\textwidth}
        \centering
        \caption{$\Theta = 0$, $D = 20$ cm}
        \includegraphics[width=\linewidth]{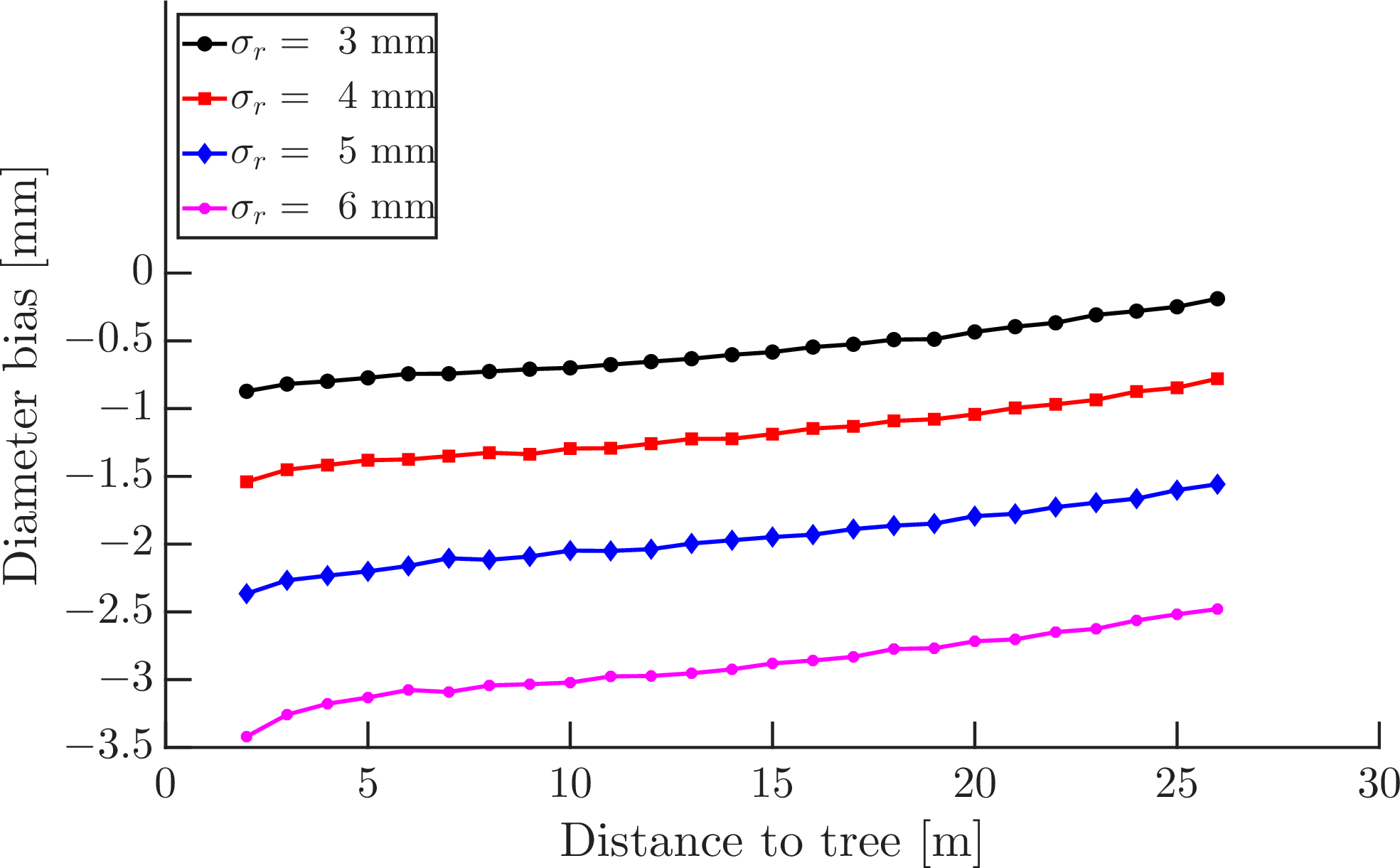}
    \end{subfigure}
    \hfill
    \begin{subfigure}{0.48\textwidth}
        \centering
        \caption{$\Theta = 0$, $d = 10$ m}
        \includegraphics[width=\linewidth]{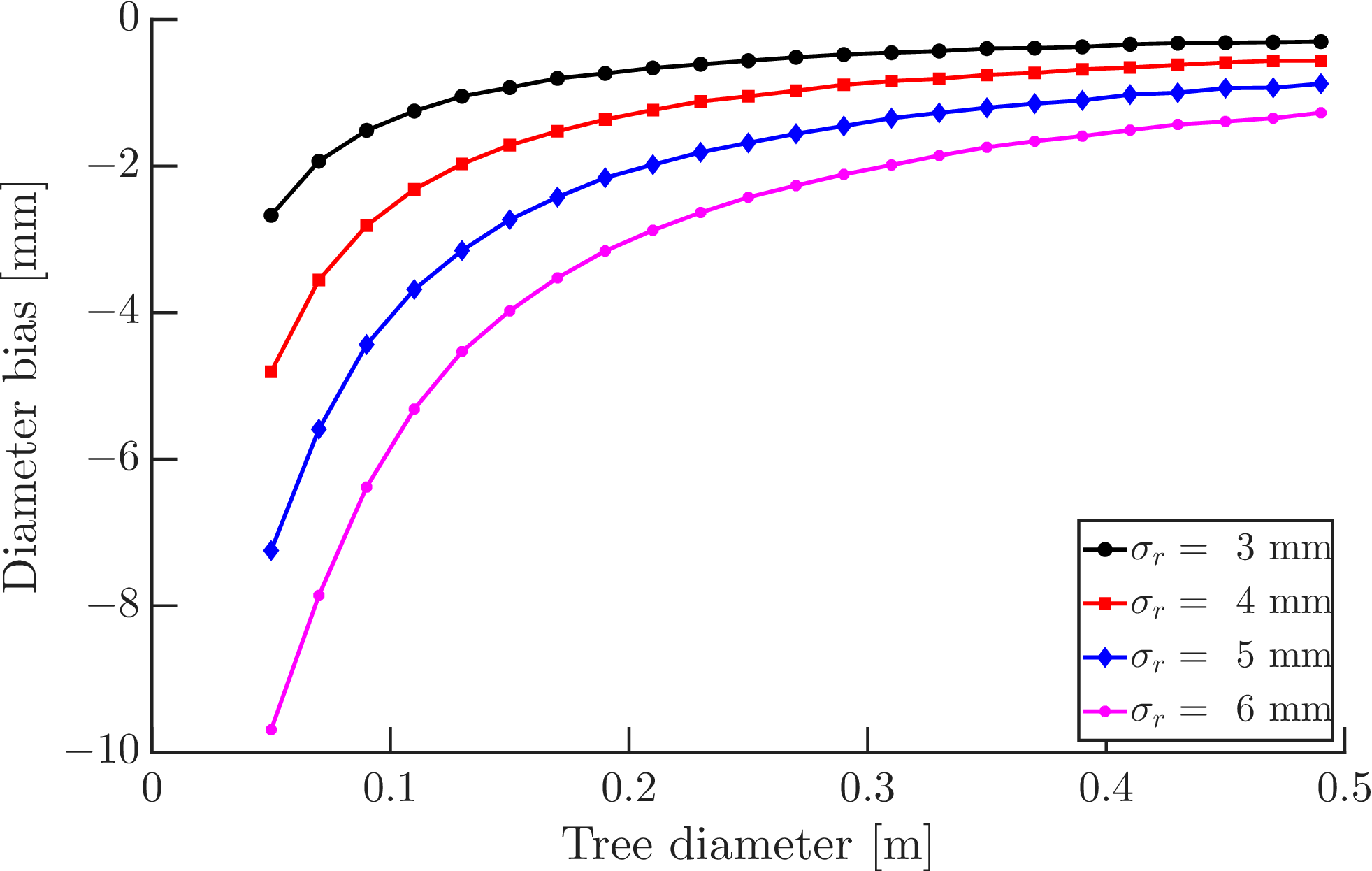}
    \end{subfigure}

    \vspace{0.5cm}

    \begin{subfigure}{0.48\textwidth}
        \centering
        \caption{Variable $\Theta$ and $\sigma_r$, $D = 20$ cm}
        \includegraphics[width=\linewidth]{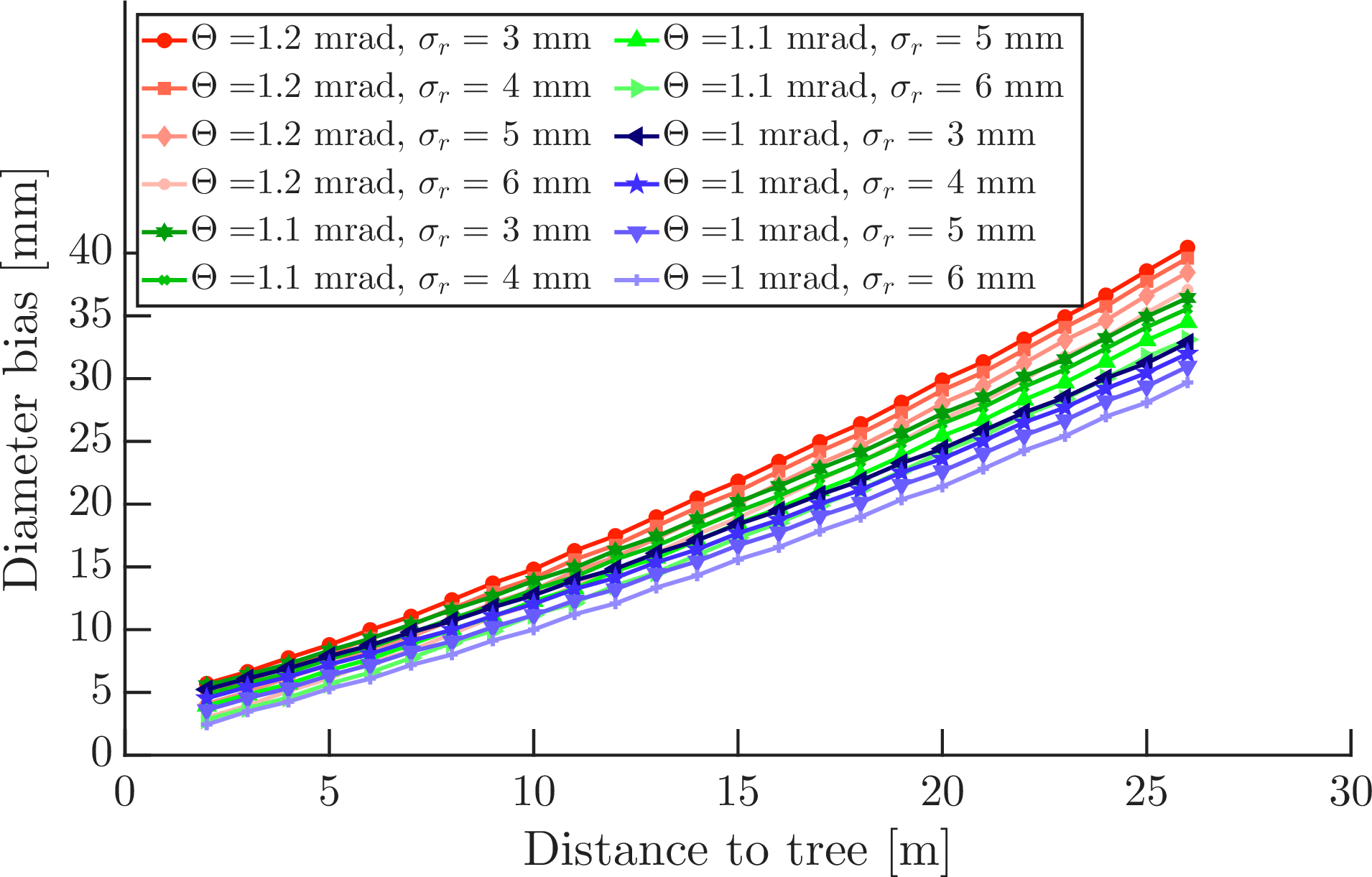}
    \end{subfigure}
    \hfill
    \begin{subfigure}{0.48\textwidth}
        \centering
        \caption{Variable $\Theta$ and $\sigma_r$, $d = 10$ m}
        \includegraphics[width=\linewidth]{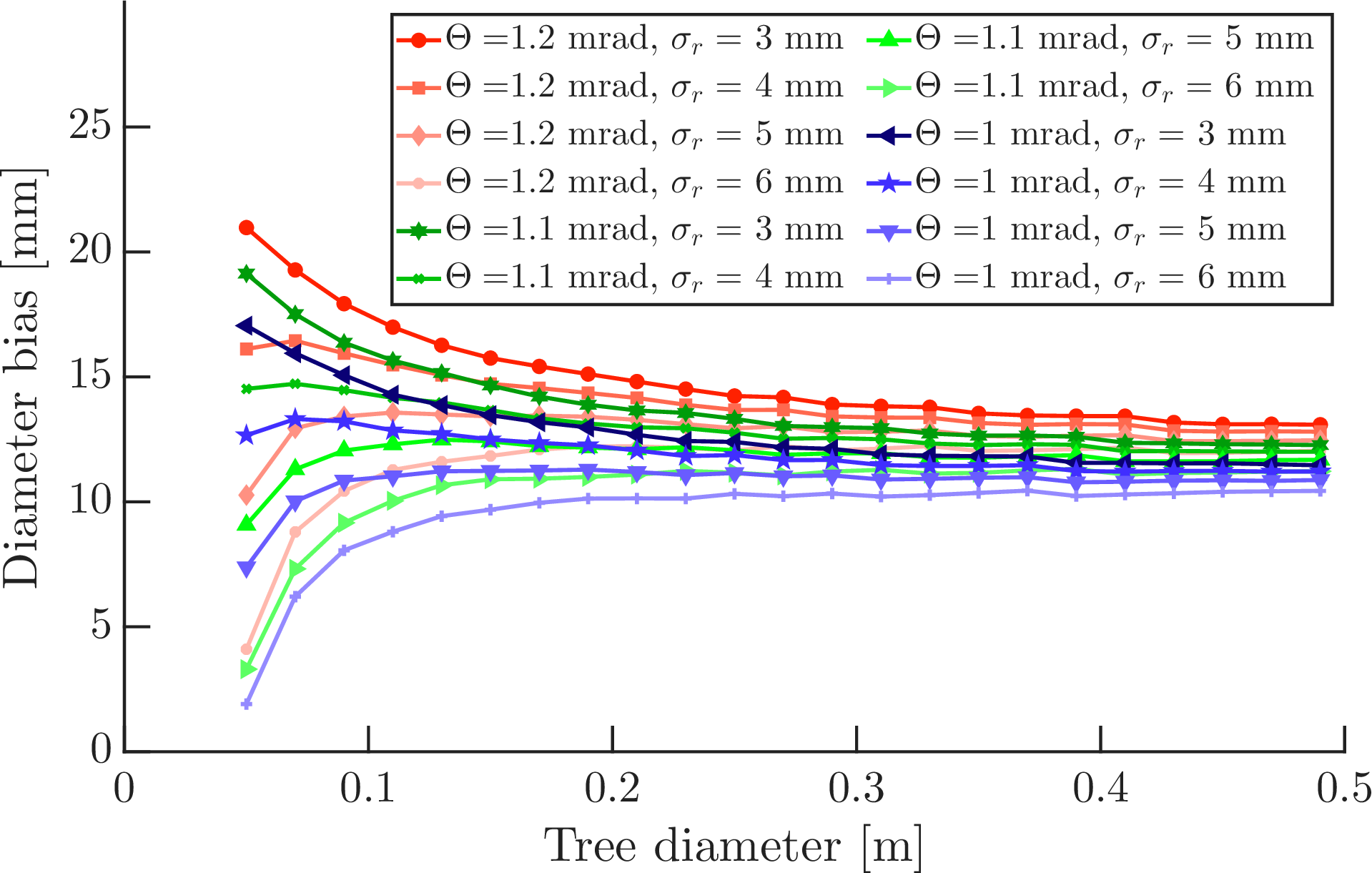}
    \end{subfigure}

    \caption{Simulated bias as a function of measurement distance ((a), (c) and (e)), and tree diameter ((b), (d) and (f)). For distance simulations the tree diameter $D$ was constant \qty{20}{\centi \metre} and the measurement distance $d$ was \qty{10}{\metre} for the diameter simulations. In (e) and (f) the divergences are grouped by color with lighter colors indicating higher ranging noise.}
    \label{fig: simulation results}
\end{figure*}

Figure~\ref{fig: simulation results} presents the main simulation results. The tree diameter bias as a function of distance to the tree is shown in Figs.~\ref{fig: simulation results}(a), (c) and (e) with the constant reference diameter of \qty{20}{\centi\metre}, and as a function of tree diameter in Figs.~\ref{fig: simulation results}(b), (d) and (f) with a constant measurement distance of \qty{10}{\metre}. Figures~\ref{fig: simulation results}(a) and (b) illustrate the impact of beam divergence $\Theta$ on diameter bias, while the random measurement noise was set to zero ($\sigma_r = \sigma_{\theta}=0$). In contrast, Figures~\ref{fig: simulation results}(c) and (d) show the dependence of diameter bias on range noise $\sigma_r$, while  the laser beam was assumed infinitely narrow. In Figs.~\ref{fig: simulation results}(e) and (f), both range noise $\sigma_r$ and beam divergence $\Theta$ were varied, while angle noise $\sigma_{\theta}$ was constant at $\sigma_{\theta} = 10^{-4}$ rad.

\begin{figure*}[htbp]
    \centering

    \begin{subfigure}{0.48\textwidth}
        \centering
        \caption{$\sigma_r = 0$, $D = 20$ cm}
        \includegraphics[width=\linewidth]{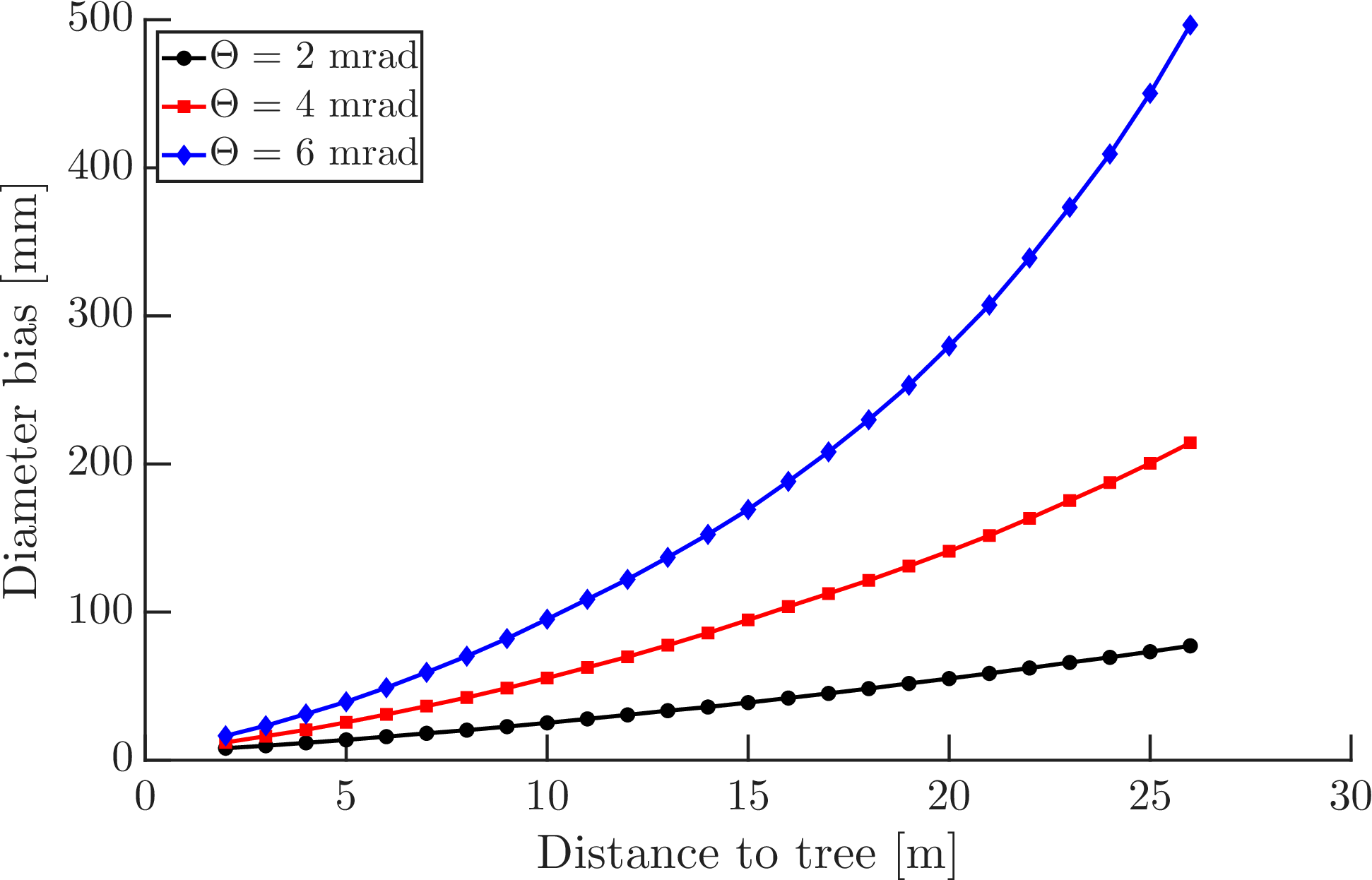}
    \end{subfigure}
    \hfill
    \begin{subfigure}{0.48\textwidth}
        \centering
        \caption{$\sigma_r = 0$, $d = 10$ m}
        \includegraphics[width=\linewidth]{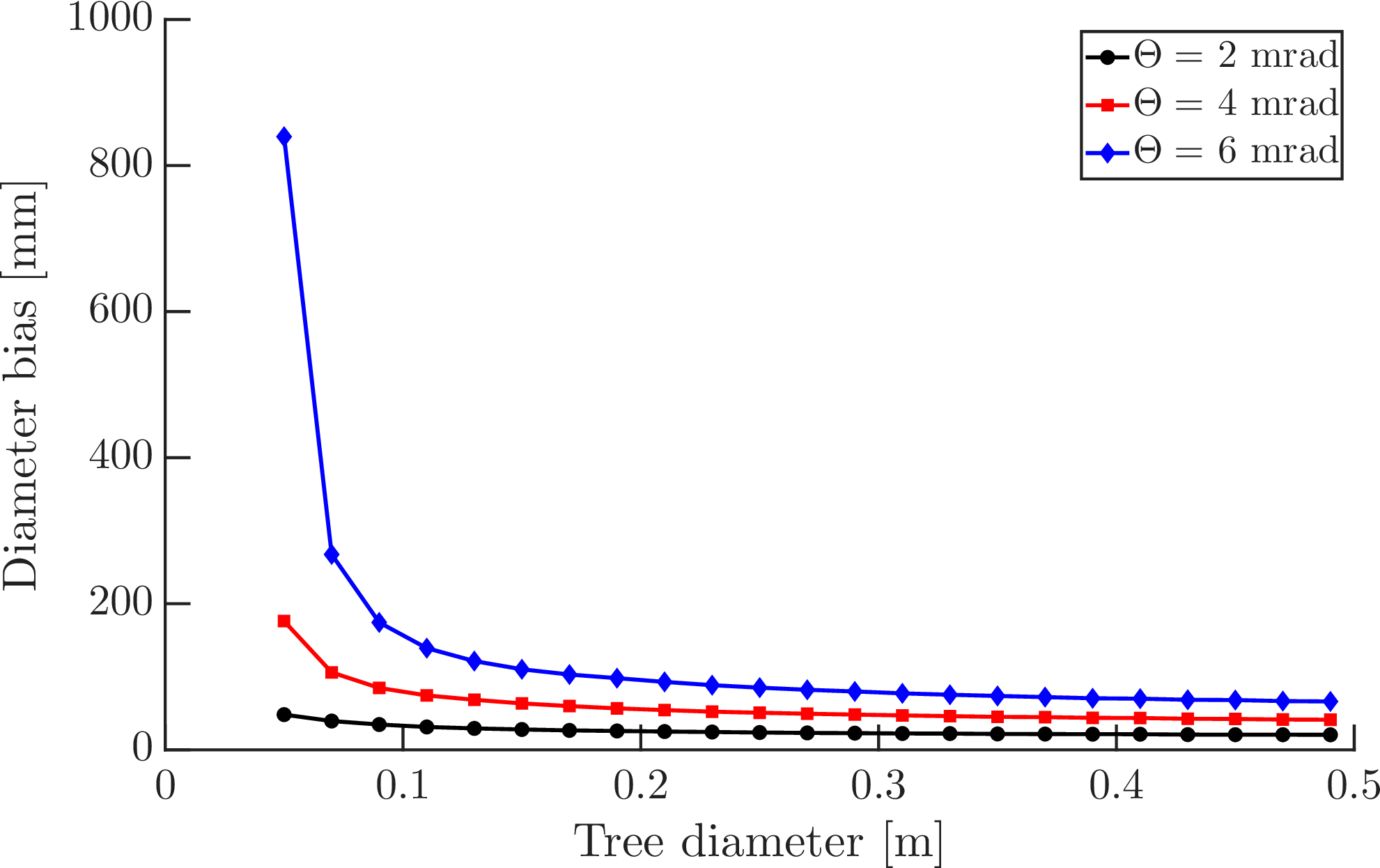}
    \end{subfigure}

    \vspace{0.5cm}

    \begin{subfigure}{0.48\textwidth}
        \centering
        \caption{$\Theta = 0$, $D = 20$ cm}
        \includegraphics[width=\linewidth]{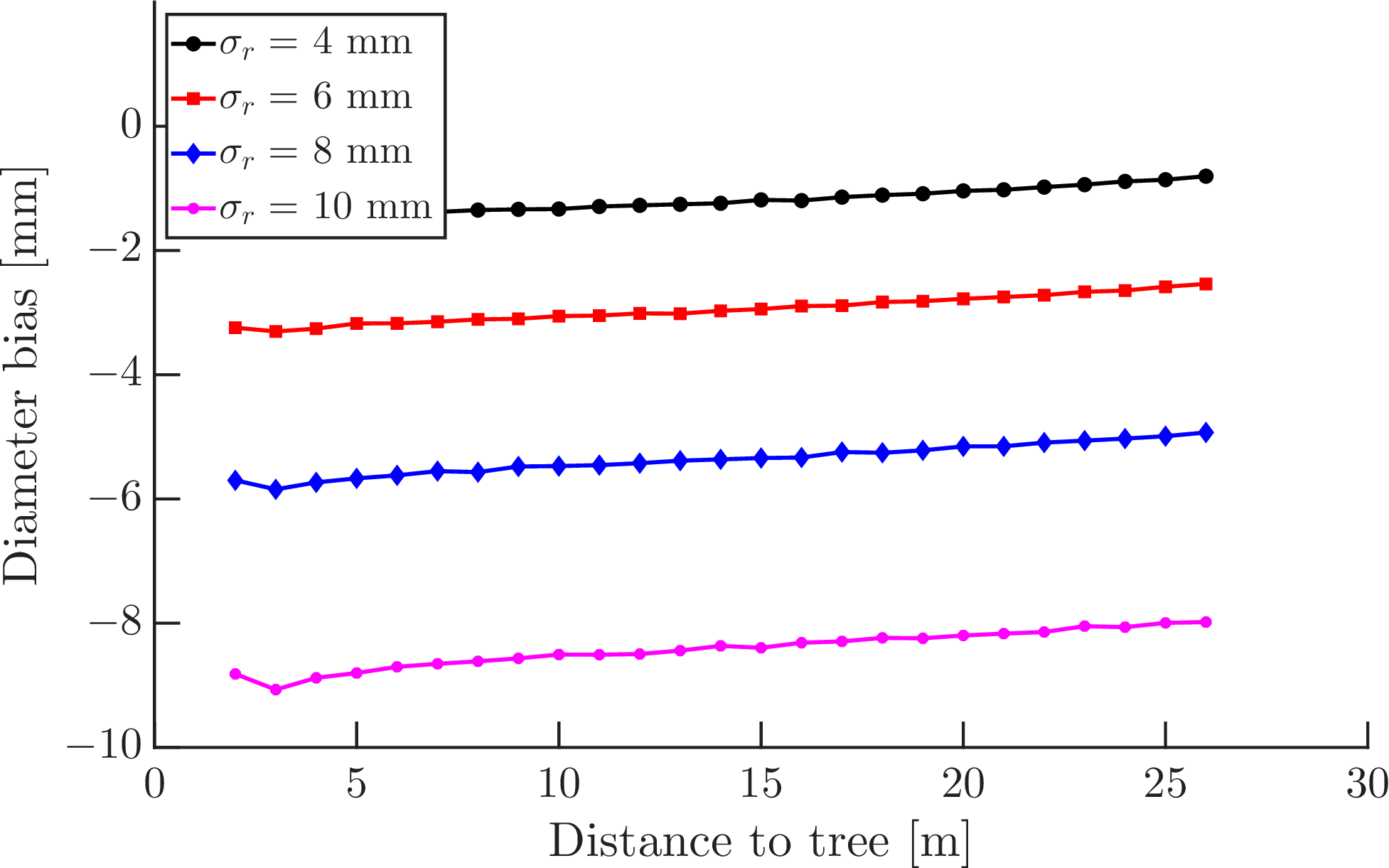}
    \end{subfigure}
    \hfill
    \begin{subfigure}{0.48\textwidth}
        \centering
        \caption{$\Theta = 0$, $d = 10$ m}
        \includegraphics[width=\linewidth]{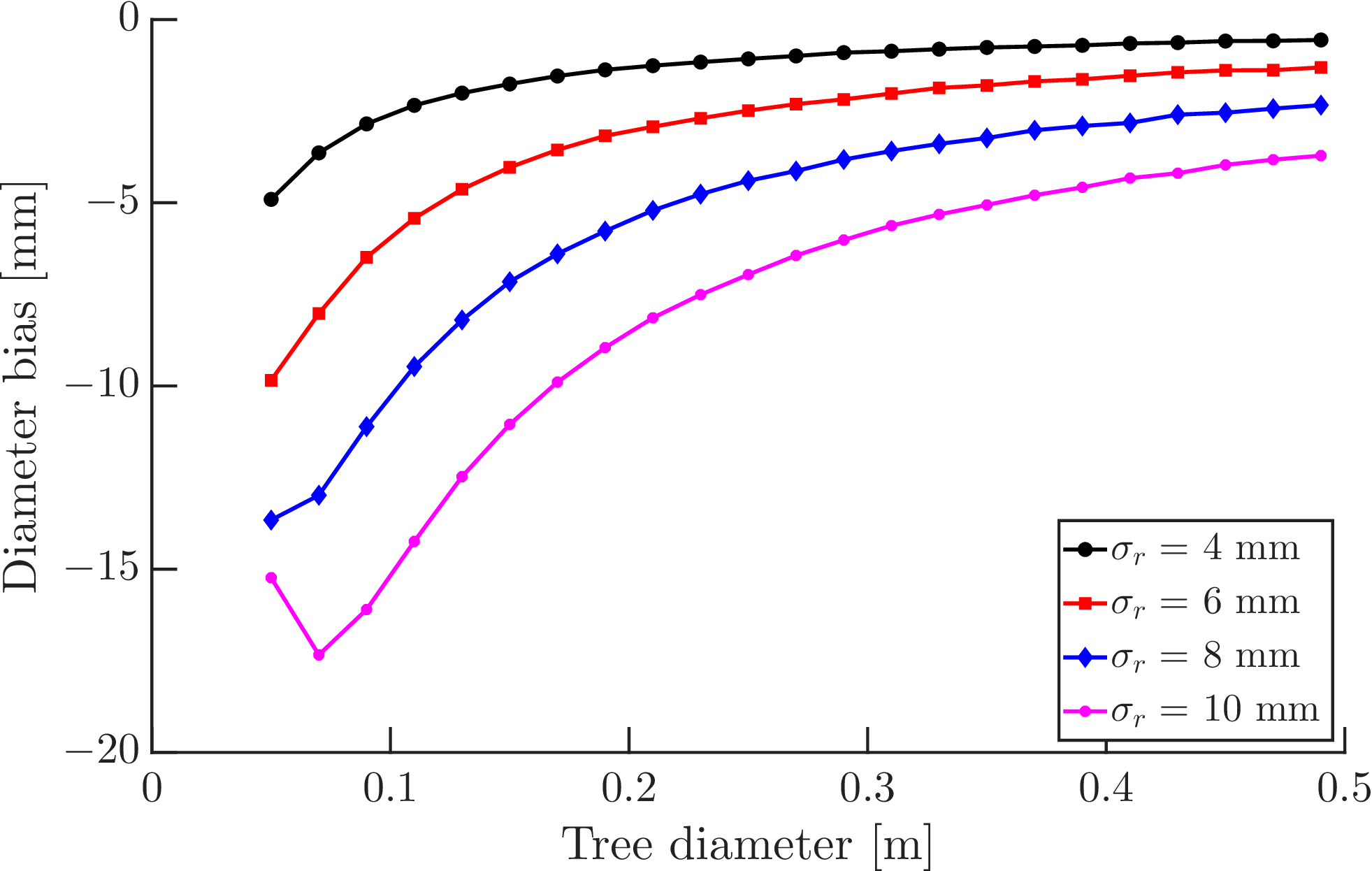}
    \end{subfigure}

    \vspace{0.5cm}

    \begin{subfigure}{0.48\textwidth}
        \centering
        \caption{Variable $\Theta$ and $\sigma_r$, $D = 20$ cm}
        \includegraphics[width=\linewidth]{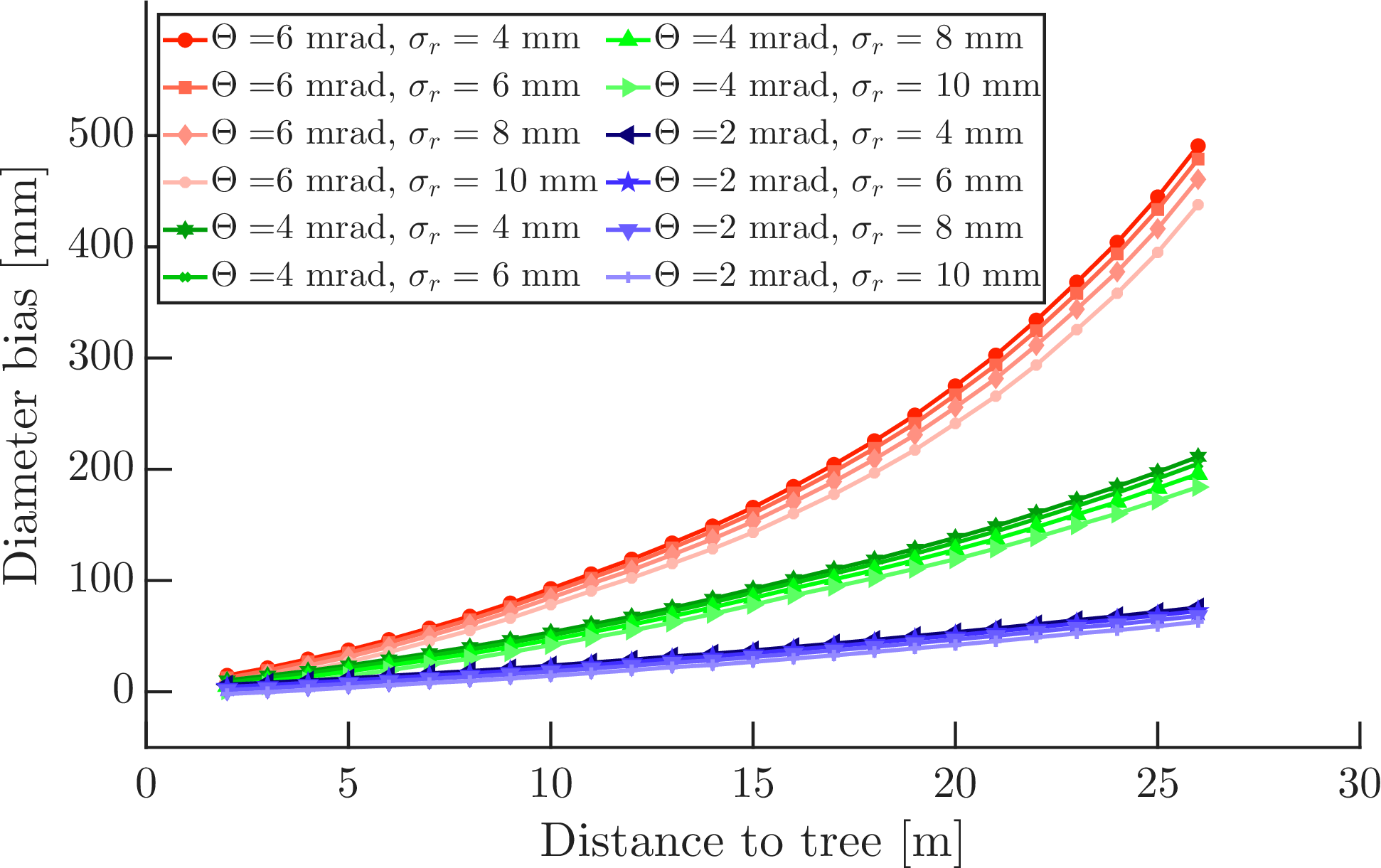}
    \end{subfigure}
    \hfill
    \begin{subfigure}{0.48\textwidth}
        \centering
        \caption{Variable $\Theta$ and $\sigma_r$, $d = 10$ m}
        \includegraphics[width=\linewidth]{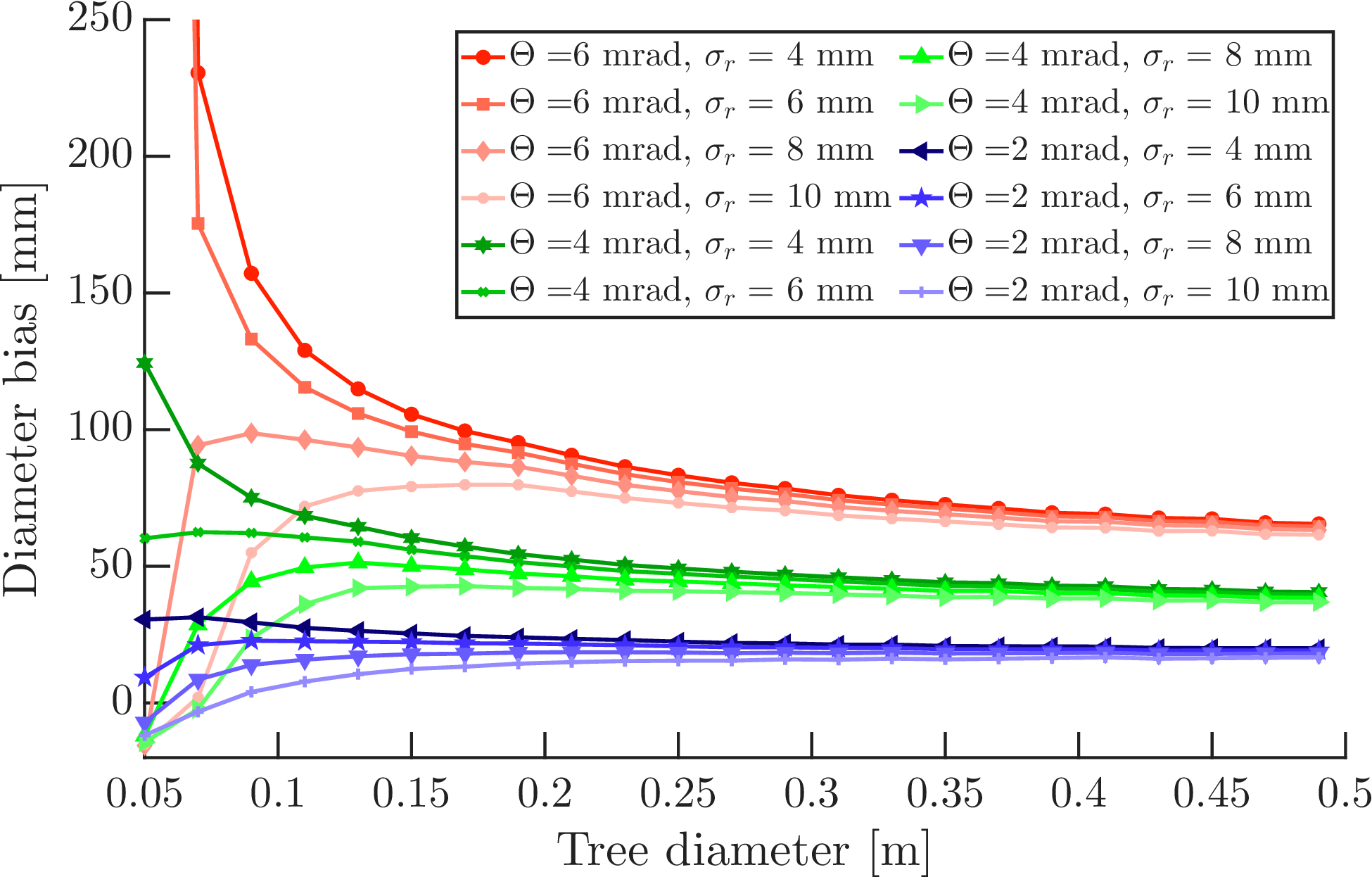}
    \end{subfigure}

    \caption{Simulated bias as a function of measurement distance ((a), (c) and (e)), and tree diameter ((b), (d) and (f)) using a large spread of simulation parameters. For distance simulations the tree diameter $D$ was constant \qty{20}{\centi \metre} and the measurement distance $d$ was \qty{10}{\metre} for the diameter simulations. In (e) and (f) the divergences are grouped by color with lighter colors indicating higher ranging noise.}
    \label{fig: simulation results large param spread}
\end{figure*}

When the finite beam width effect was combined with stochastic noise, the simulation reproduced all the different bias behaviors reported in the literature: Figures~\ref{fig: simulation results}(e) and \ref{fig: simulation results large param spread}(e) show both near linear- and near quadratic bias growth with respect to the measurement distance and Figs.~\ref{fig: simulation results}(f)  and \ref{fig: simulation results large param spread}(f) show both decreasing and increasing bias as a function of tree diameter---depending on the simulation parameters. The simulations also agree on the common finding that the magnitude of bias---both positive and negative---is significantly larger for small trees with DBH smaller than \qtyrange[range-phrase = --]{10}{15}{\centi\metre}~\citep{ryding_assessing_2015, gollob_forest_2020, kuzelka_inventory_2022}. Furthermore, while the simulations produced mostly positive diameter biases, negative biases were also observed, mainly for small trees measured from close range (Fig.~\ref{fig: simulation results large param spread}(f)) and when using $t_{\mathrm{thresh}} = 0.75$ in Section~\ref{sec: cylinder measurements} (the black cylinder measurement). Sensitivity analysis of the lidar simulations to $t_{\mathrm{thresh}}$ is presented in Appendix \ref{app: trigger threshold}. 

\subsubsection{Cylinder Measurements Under Laboratory Conditions}
\label{sec: cylinder measurements}

\begin{figure*}[htbp]
    \centering

    \begin{subfigure}{0.48\textwidth}
        \centering
        \caption{Brown}
        \includegraphics[width=\linewidth]{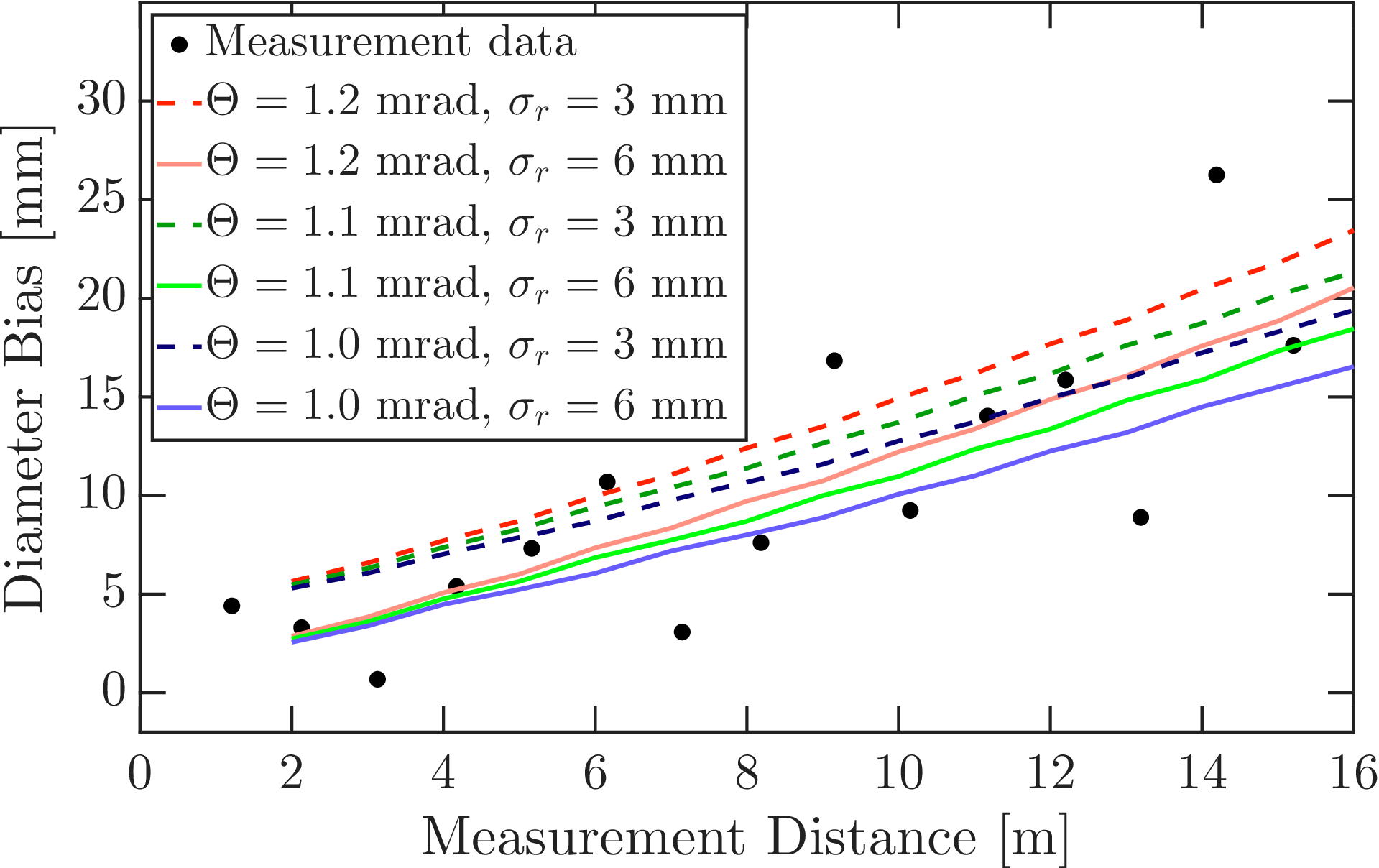}
    \end{subfigure}
    \hfill
    \begin{subfigure}{0.48\textwidth}
        \centering
        \caption{White}
        \includegraphics[width=\linewidth]{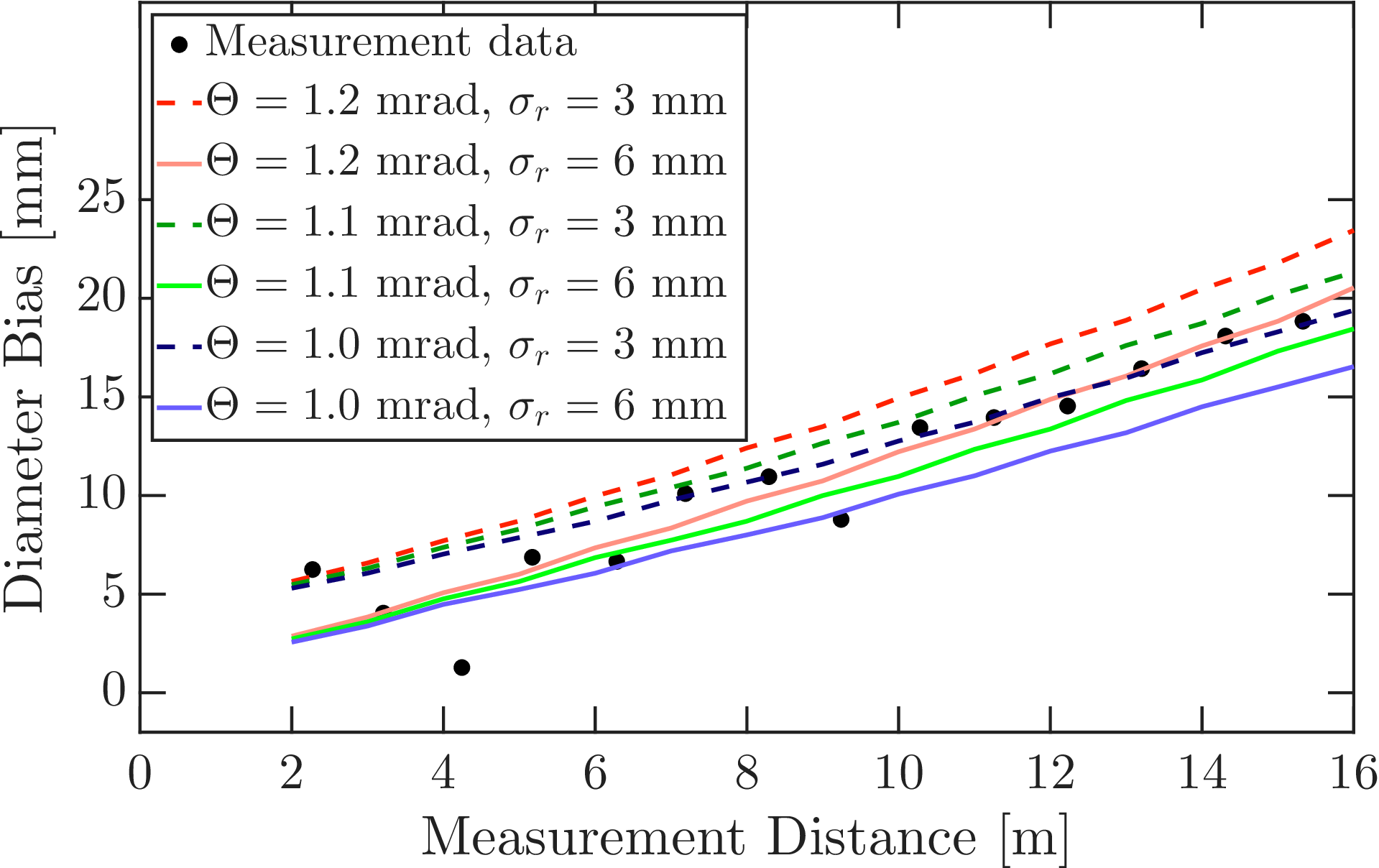}
    \end{subfigure}
    
   \begin{subfigure}{0.9\textwidth}
        \centering
        \caption{Black}
        \includegraphics[width=\linewidth]{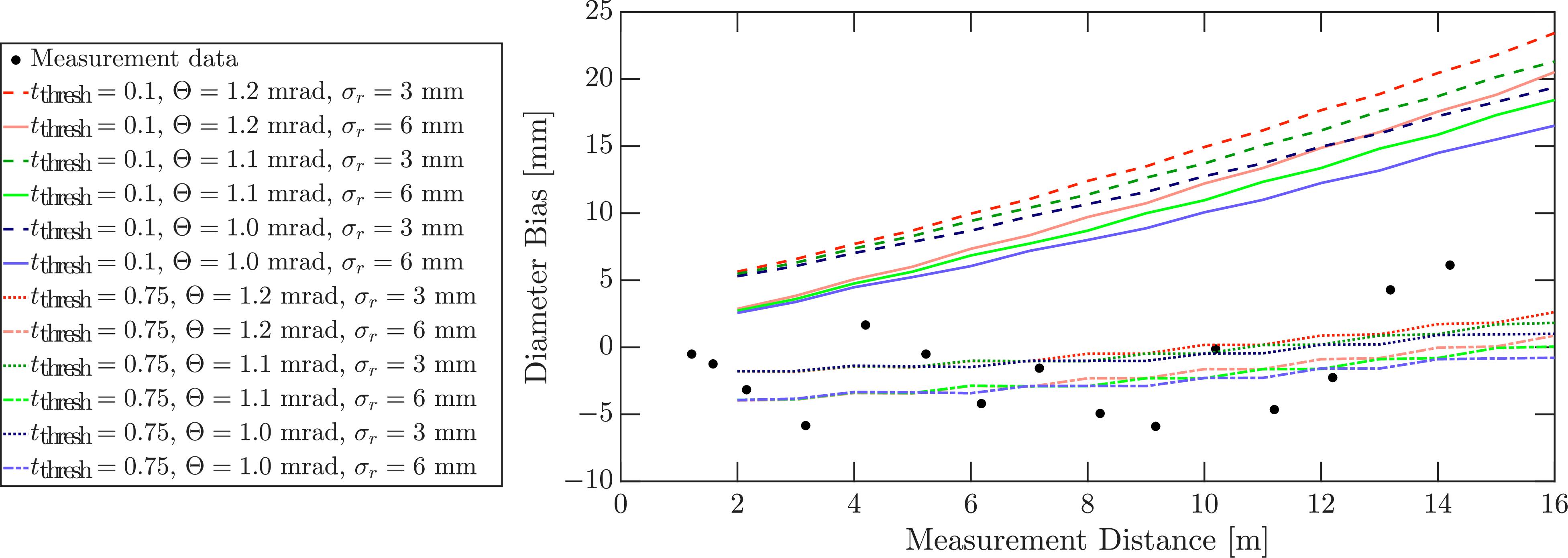}
    \end{subfigure}
    \caption{Measurements of the distance-dependent bias using the artificial cylinder. The cylinder was covered with a (a) brown, (b) white and (c) black thin cardboard. The simulations with extreme simulation parameter values (see Fig.~\ref{fig: simulation results}) are also presented as colored lines. In addition, simulations with ($t_\mathrm{thresh} = 0.75$ are presented for the black cardboard.}
    \label{fig: pipe results}
\end{figure*}

Figure~\ref{fig: pipe results} shows the diameter bias as a function of measurement distance from the \qty{210}{\milli\metre} diameter cylinder, covered in (a) brown, (b) white and (c) black paper. The figure also show selected bias curves simulated with values close to the manufacturer-declared values and $t_\mathrm{thresh} = 0.1$, the standard value used thorough this study. In addition, $t_\mathrm{thresh} = 0.75$ was used for the black paper to simulate its significantly lower reflectivity compared to the brown and white paper, which have reflectivity values close to tree bark. The simulated lines follow the measurement well in (a) and (b) (brown and white paper, respectively). Bias increases linearly in both measurements, although the measurement noise is quite high despite the ideal measurement environment. Surprisingly, the noise was higher for brown paper than the white paper. The black paper (Fig.~\ref{fig: pipe results}(c)) behaved noticeably different, as the bias stayed at or bellow zero up until approximately \qty{13}{\metre}. This behavior was successfully simulated by increasing the trigger threshold to 0.75, which simulates lower proportion of light reflecting back from the target, as discussed in Section~\ref{sec: simulations}.

\subsubsection{Experiments on Forest Sample Plots}
\label{sec: results forest plots}

\begin{figure*}[htbp]
    \centering

    \begin{subfigure}{0.48\textwidth}
        \centering
        \caption{}
        \includegraphics[width=\linewidth]{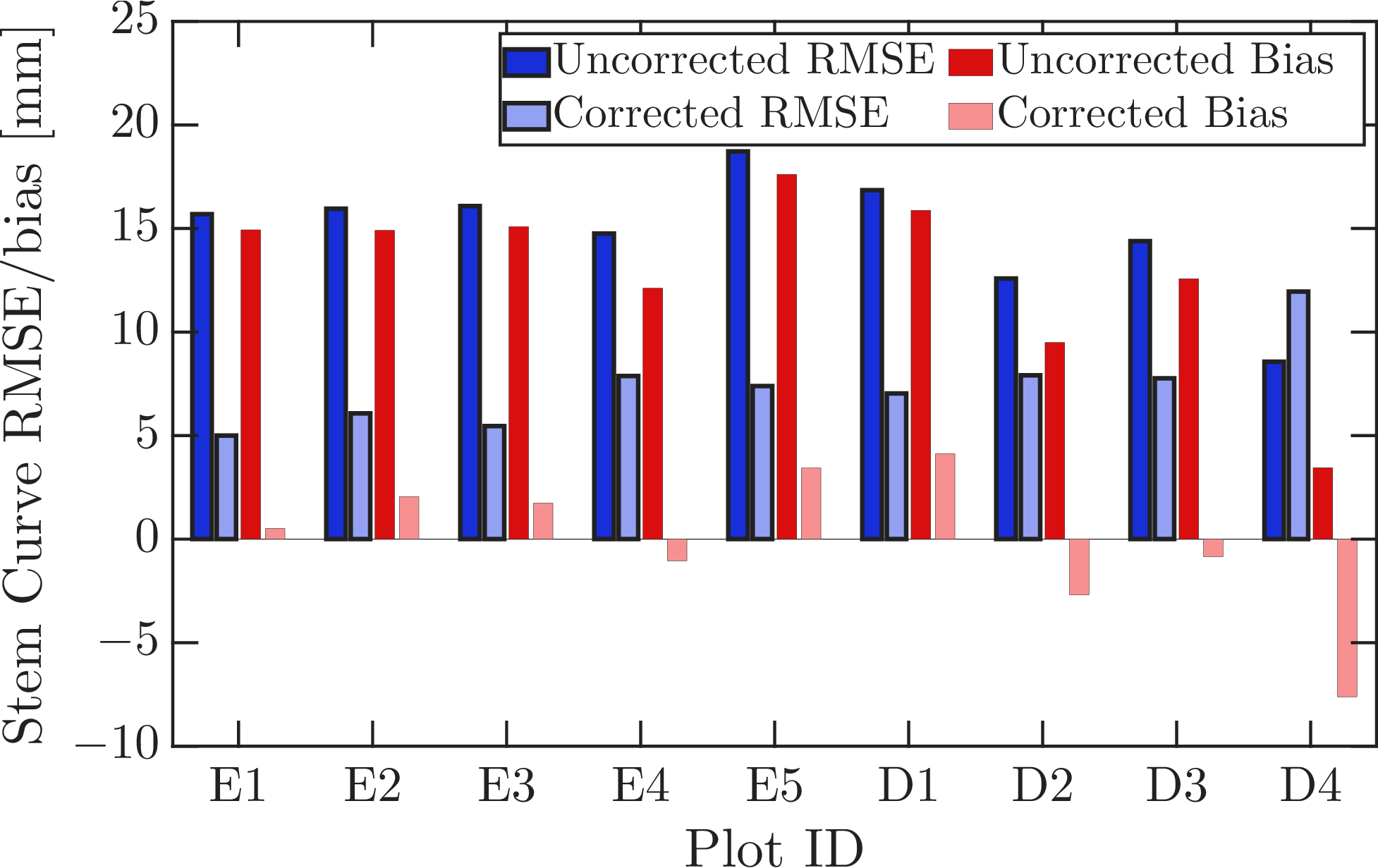}
    \end{subfigure}
    \hfill
    \begin{subfigure}{0.48\textwidth}
        \centering
        \caption{}
        \includegraphics[width=\linewidth]{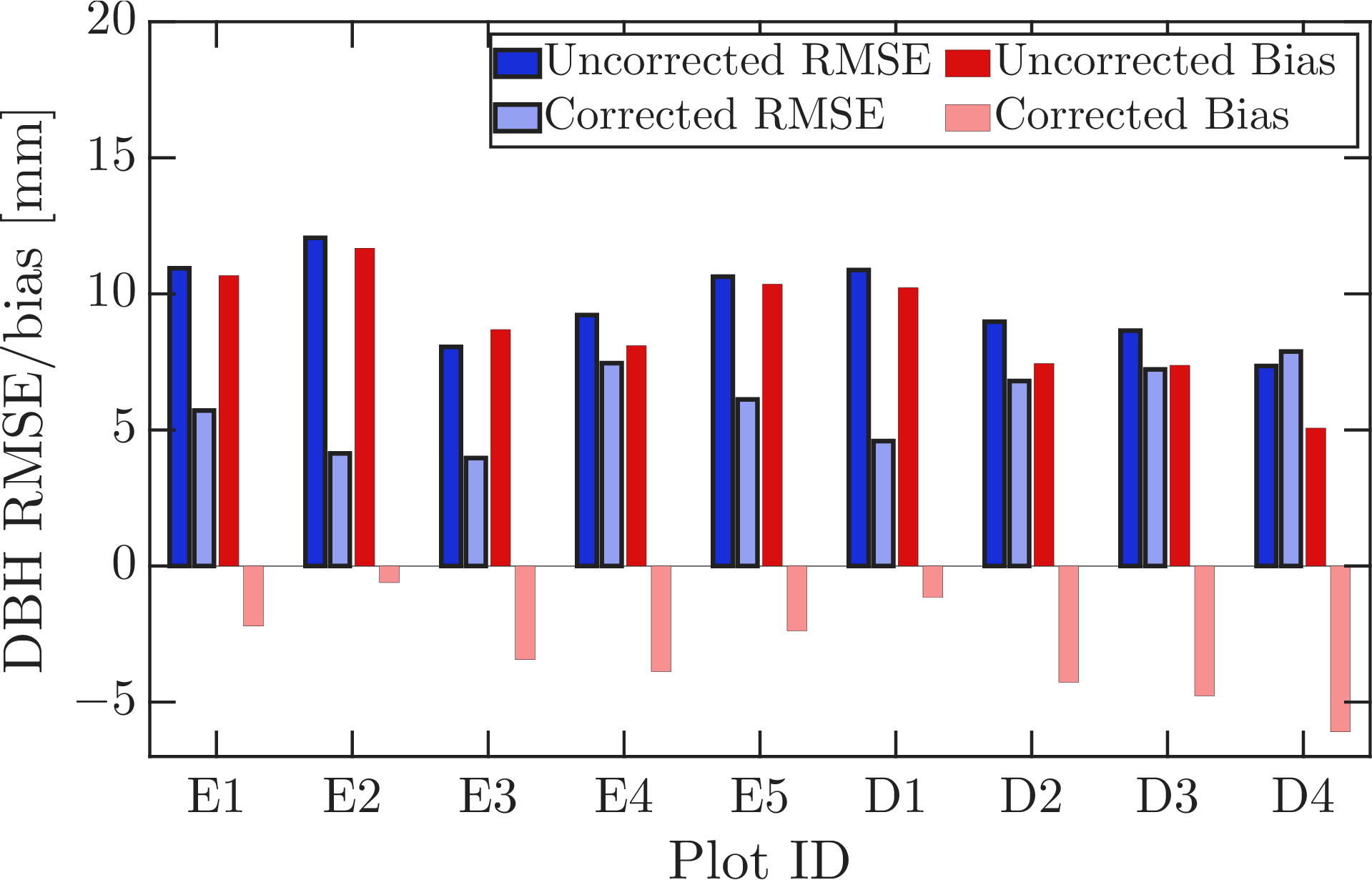}
    \end{subfigure}

    \vspace{0.5cm}

    \begin{subfigure}{0.48\textwidth}
        \centering
        \caption{}
        \includegraphics[width=\linewidth]{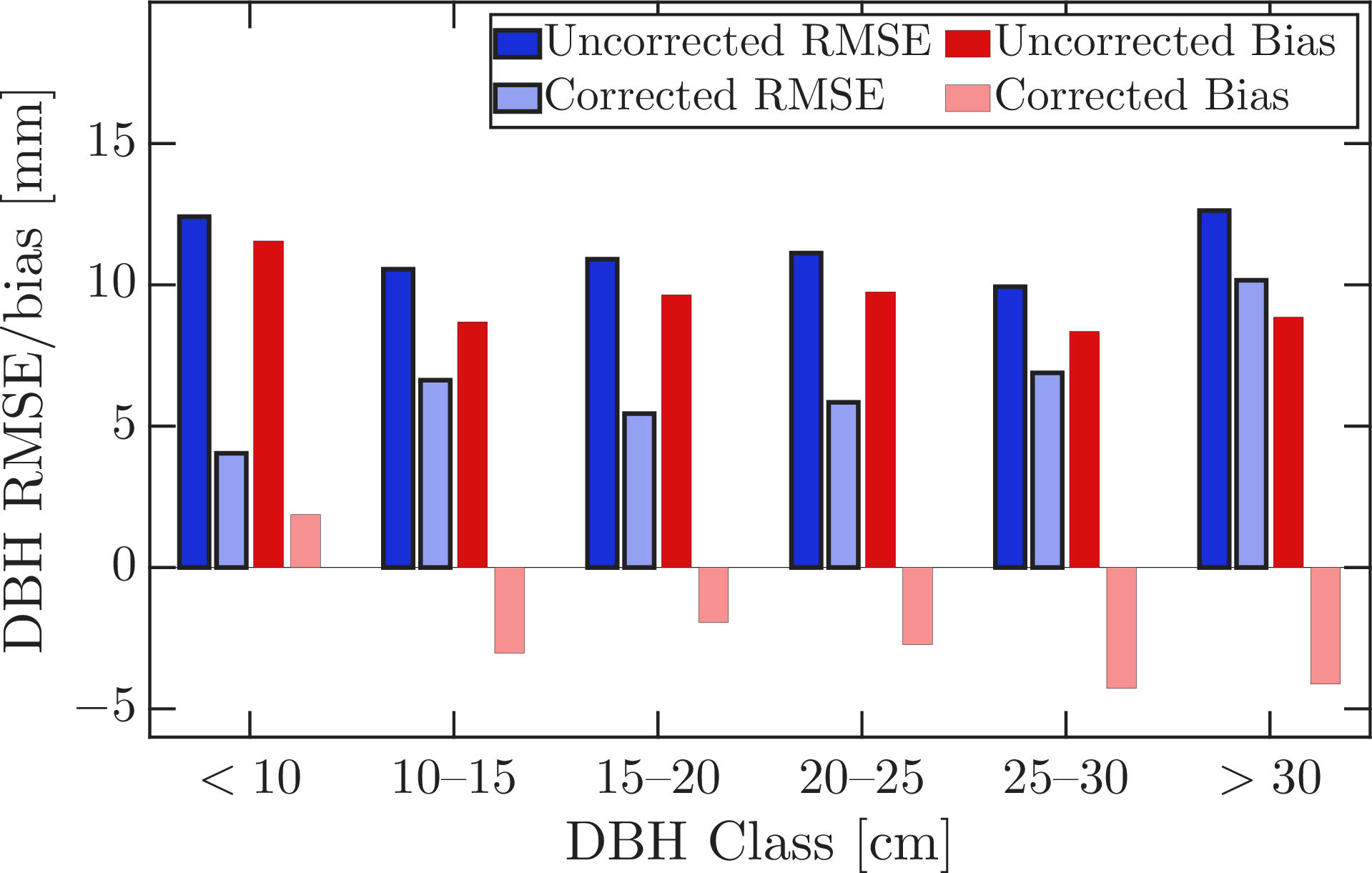}
    \end{subfigure}
    \hfill
    \begin{subfigure}{0.48\textwidth}
        \centering
        \caption{}
        \includegraphics[width=\linewidth]{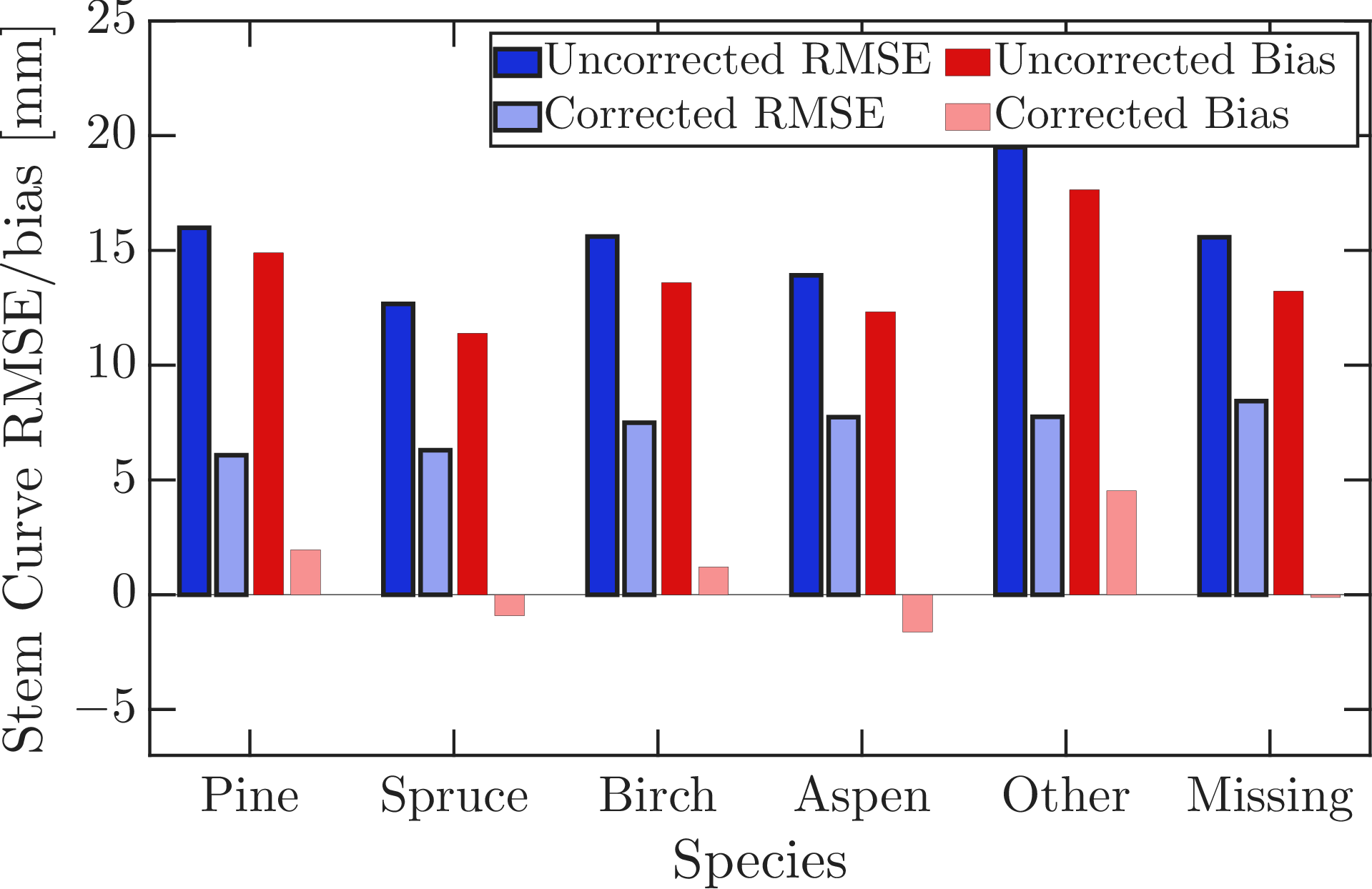}
    \end{subfigure}

    \vspace{0.5cm}

    \caption{Results from the forest plots. (a) and (b) present the plot-wise RMSE and bias for stem curve and DBH, respectively. (c) shows the DBH error metrics of different DBH classes and (d) shows stem curve error metrics for different tree species. Note that the results from the anomalous plot D4 was removed from (c) and (d) to mainly prevent it from skewing the results of spruce and aspen in (d), as discussed in Sec.~\ref{sec: results forest plots}.}
    \label{fig: forest plots main results}
\end{figure*}

Figure~\ref{fig: forest plots main results} presents the plot-, DBH class- and species-wise results from the Evo test site for the datasets collected using the Faro Orbis mobile laser scanner. We present both the results acquired with the method introduced in \citet{hyyppa_under-canopy_2020} and its bias-compensated extension developed in this study and presented in Section~\ref{sec: implementation}. Figures~\ref{fig: forest plots main results}(a) and (b) present the RMSE (thick black outline) and bias of stem curve and DBH, respectively for each sample plot. The "E" labels represent easy plots, and the "D" labels difficult ones. For uncorrected diameters, the RMSE and bias (Eqs.~\eqref{eq: RMSE} and \eqref{eq: bias}) of stem curve were \qtyrange[range-phrase = --]{14}{17}{ \milli\metre } and \qtyrange[range-phrase = --]{9}{16}{ \milli\metre}, respectively,  excluding the anomalous plot D4. After the bias correction, the RMSE reduced to \qtyrange[range-phrase = --]{5}{7}{ \milli\metre } and the bias to \qtyrange[range-phrase = --]{-2}{4}{ \milli\metre }. RMSE and bias for uncorrected DBH was \qtyrange[range-phrase = --]{8}{12}{ \milli\metre } for both metrics and \qtyrange[range-phrase = --]{3}{7}{ \milli\metre } and \qtyrange[range-phrase = --]{-4}{0}{ \milli\metre } for the corrected DBHs. 

The simulation-based bias compensation provided significant improvement in the RMSE of stem curve estimates, even though the RMSE of the uncompensated method was already comparable to the state-of-the-art. The bias compensation allows MLS to provide highly accurate---close to TLS level---diameter measurements from trees/sample plots, with less time spent on the sample plots compared to TLS or manual measurements. The bias compensated RMSE for the stem curves of \qtyrange[range-phrase = --]{5}{10}{\milli\metre} is close to the accuracy limit achievable by approximating stems as circles, due to the bark roughness and out-of-roundness of the stems. 

The plot D4 was anomalous and had a very low stem curve bias of \qty{3}{\milli\metre} without correction. This resulted in the correction to overshoot the diameter reduction down to approximately \qty{-6}{\milli\metre}. A closer look found no explanation for this; the diameter distribution of trees was similar to other difficult plots, and the main tree species of the plot---spruce and aspen---were also found in other plots, where no such effect was detected. The scanning conditions were similar in both the TLS and MLS measurements and they were temporally separated by only 2 days---less than most plots. Furthermore, the smaller bias was found to be systematic and not a result of outliers. Therefore, we excluded the plot D4 from results other than Fig.~\ref{fig: forest plots main results}(a) and (b), as the anomalous results would skew the results, especially for spruce and aspen, which produced similar results to other tree species on other plots.

Figures~\ref{fig: forest plots main results}(c) and (d) present the combined results from all plots. (c) presents the DBH RMSE and bias for each DBH class and (d) the stem curve RMSE and bias for each main tree species. The species are Pine (\textit{Pinus sylvestris}), spruce (\textit{Picea abies}), birch (\textit{Betula pendula} and \textit{Betula pubescens}) and aspen (\textit{Populus Tremula}). Other deciduous species were placed in the "other" category, and the trees with no species information into "missing" category. We found no significant difference between the main tree species and the error values were roughly the same as for plot-wise results presented in Fig.~\ref{fig: forest plots main results}(a). The RMSE and bias were somewhat higher for category "other" trees, although this is most likely caused by the fact that the other deciduous trees were in general quite small and often had curved and out-of-round stems.

\begin{figure}[htbp]
    \begin{subfigure}{0.49\textwidth}
        \centering
        \caption{}
        \includegraphics[width=\linewidth]{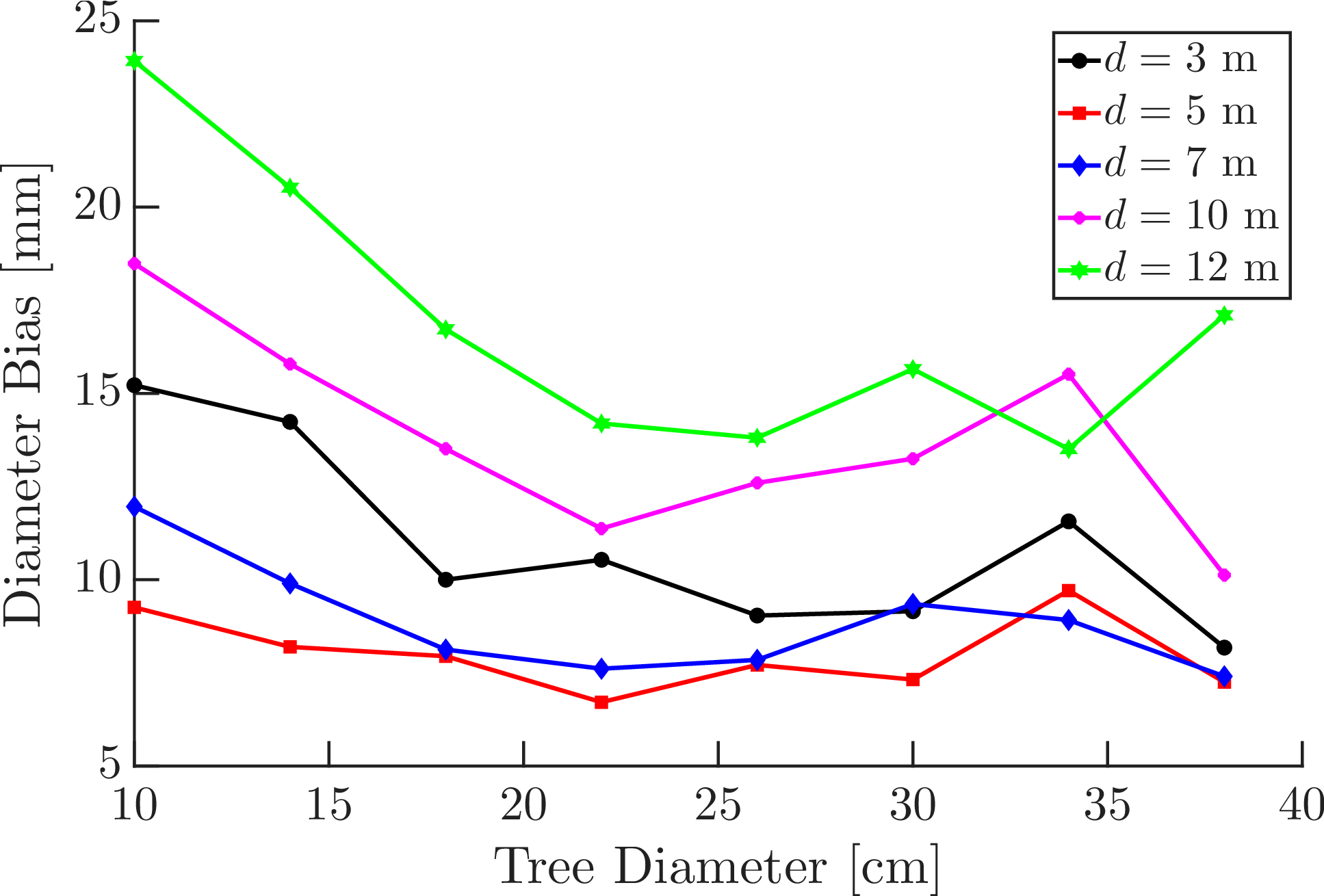}
    \end{subfigure}
    \hfill
    \begin{subfigure}{0.49\textwidth}
        \centering
        \caption{}
        \includegraphics[width=\linewidth]{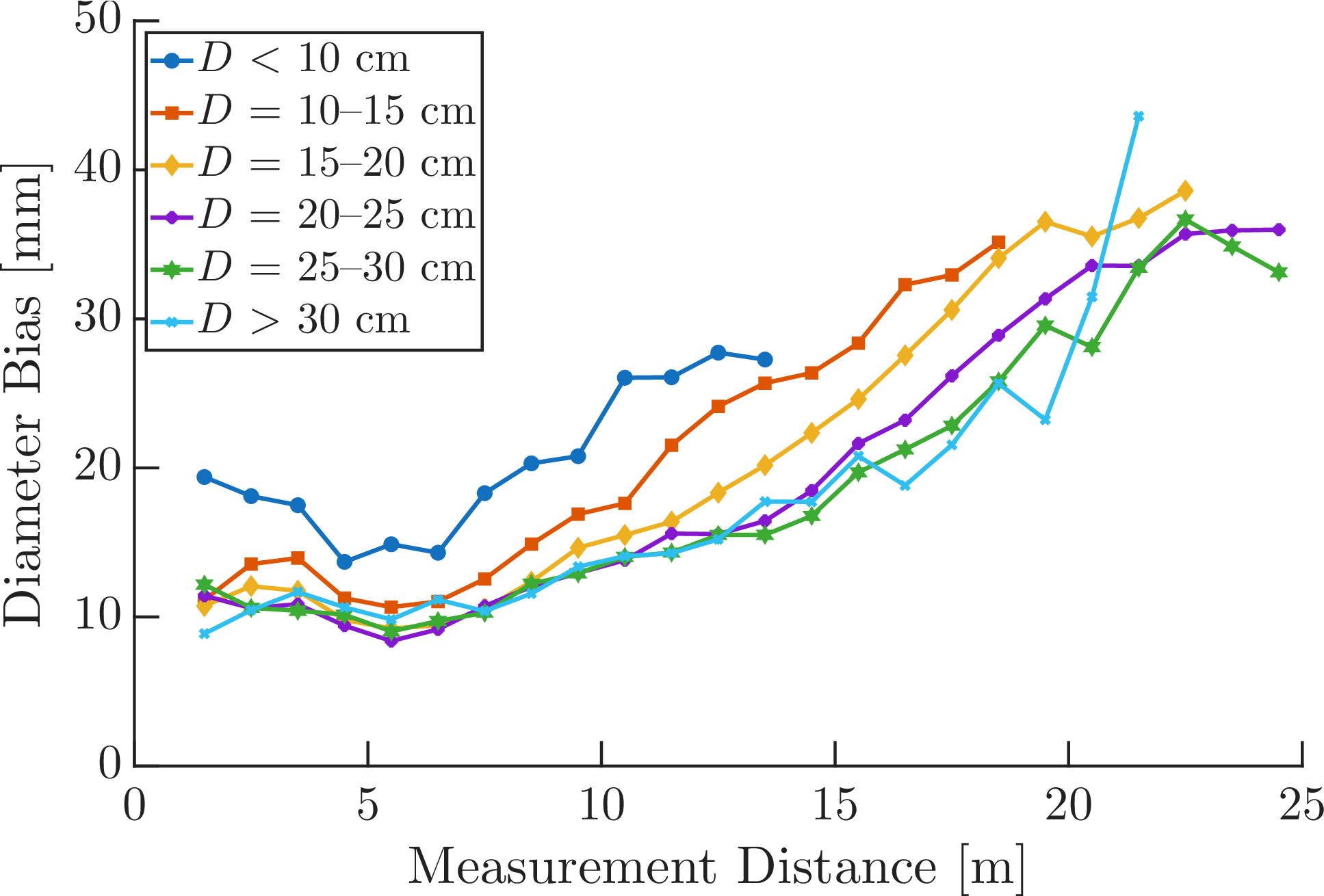}
    \end{subfigure}
    \caption{(a) Diameter bias as a function of tree diameter for a fixed measurement distance and a fixed measurement height of \qtyrange[range-phrase = --]{1.5}{2.5}{\metre} (without the bias compensation). (b) Diameter bias as a function of measurement distance for fixed diameter classes. Measurement points with less than 50 arcs were removed from the results.}
\label{fig: single distance}
\end{figure}

Figure~\ref{fig: single distance}(a) shows the tree diameter bias as a function of tree diameter for the fixed measurement distances and a fixed measurement height without the bias compensation. The $x$-axis shows the center of the diameter bins. The diameter bias tends to decrease as the tree diameter increases, and varies from \qty{9}{\milli\metre} to \qty{24}{\milli\metre} for \qtyrange[range-phrase = --]{8}{12}{\centi\metre} trees and from \qty{7}{\milli\metre} to \qty{18}{\milli\metre} for \qtyrange[range-phrase = --]{36}{40}{\centi\metre} trees. The observed diameter dependency is greater for far measurements but overall quite small---at maximum the bias decreased by approximately \qty{10}{mm} between the trees with diameter of \qty{10}{cm} and \qty{38}{cm}. 

The slightly decreasing bias as a function of the tree diameter, that leveled off to a constant at a diameter of approximately \qty{20}{cm}, was predicted by the simulations with suitable parameters (see Fig.~\ref{fig: simulation results}(f)). However, as mentioned previously, the simulated bias for small trees was highly sensitive to the simulation parameters. This sensitivity is a difficult problem to solve, as even with no measurement noise, the bark of the tree essentially causes noise that is dependent on not only the tree species, but the individual tree as well. However, we found no dependency between bias and tree species for the main species (see Fig.~\ref{fig: forest plots main results}(d)). 

Figure~\ref{fig: single distance}(b) presents analogous results, where the diameter bias was calculated as a function of measurement distance for fixed tree diameter ranges based on the reference diameter. The bias remained constant at around \qty{10}{mm} for trees larger than \qty{10}{cm}, until measurement distances of \qtyrange[range-phrase = --]{6}{7}{\metre}, after which the bias began to increase roughly linearly. On average, smaller trees exhibit slightly larger bias, but the effect of the tree diameter on the distance dependence of bias is small. Trees smaller than \qty{10}{cm} in diameter behave similarly, but at slightly higher overall bias. The very small decrease in bias from \qty{1}{\metre} to \qty{5}{\metre} is slightly more pronounced, suggesting that the effect may be genuine, rather than an artifact of the random measurement noise. However, the simulations did not predict this behavior, as the simulated bias increased throughout the entire range of simulated measurement distances. It is possible that this is due how Faro Orbis handles the range-walk (measurement error due to the change in the return intensity as a function of measurement distance). The (approximately) linear increase of bias at greater measurement distances than \qtyrange[range-phrase = --]{6}{7}{\metre} was predicted by the the simulations, and is in agreement with recent research \citep{kuzelka_noise_2024, muhojoki_benchmarking_2024}.

\subsection{Limitations and Further Discussion}

Although the simulations captured the diameter bias trends observed in forest sample plot measurements of this study and the past literature, the model is still incomplete. Lidar specifications, such as beam divergence and ranging noise are not always given, and they may depend on e.g.\@ measurement conditions. However, most importantly, the signal triggering and handling of range walk is typically not openly documented. Furthermore, the elliptical beam was not simulated for simplicity, as the beam orientation with respect to the stem, even within as single arc, is not constant as the scanner rotates. The current implementation required precomputed bias tables, but more computationally efficient simulation would allow for running the simulation for individual arcs or even points, and thus potentially improving the bias compensation. 

In addition, the out-of-roundness of tree stems and bark characteristics limit the potential of the simulations, especially for small trees with diameter less than \qty{10}{cm}. Even if the properties of the scanner are known, the bark depth can be similar or even greater then the noise level of lidar, potentially altering the bias-diameter curve drastically, which is further exacerbated by possible out-of-roundness of the tree stem.

One unexplored source of error and possible explanation for conflicting bias reports in the literature was the effect of arc-based diameter estimation. Using the entire point cloud for circle fitting can be more robust, as there are more points to fit the circle to, and the point cloud covers (most of the time) the stem section simultaneously from all sides. This can also reduce the effect of the edge points, i.e.\@ the points created by a partial hit of the laser beam to the edge of the stem. Therefore, the bias and random error can behave differently. The downside of not using the temporal segmentation, that was used in this study, is that any positioning errors will affect the stem geometry measurements and arc-based bias compensation methods do not work, as the measurement distance to the circle cannot be determined. Using the entire point cloud at once could potentially work with a point-by-point bias compensation method, but its development is left to future work.

\section{Conclusions}   
\label{sec: conclusions}

We investigated the effect of beam divergence and measurement noise on the estimation of tree diameter from lidar-generated point clouds. Using simulations, we found that longer measurement distances typically cause overestimation of the diameter, while high heteroscedasticity (higher angular precision than ranging precision) causes underestimation. The combined effect of these factors led to an overall overestimation of stem diameter under simulation conditions representative of state-of-the-art laser scanners and typical boreal tree reflectivity. However, small variations in the simulation parameters, namely the beam divergence and ranging noise, yielded qualitatively different behaviors, explaining the contradictory findings reported in previous studies as these parameters also vary between different laser scanning systems.

We found that for trees larger than \qty{10}{\centi\metre}, the simulations are robust with respect to the parameters, and therefore highly accurate bias correction is possible. However, for smaller trees, the bias-diameter curves are sensitive to the parameter values, which highlights the precision limitations of conventional laser scanning systems for measuring the geometry of small trees. 

Based on our findings, we developed a calibration-free bias compensation method relying on simulations and  demonstrated its potential in measurements on forest sample plots using a Faro Orbis MLS system benchmarked against a TLS reference. The method significantly reduced the RMSE of stem curve estimates to sub-centimeter level by removing most of the bias. Thus, the bias compensation method enables MLS systems to provide highly accurate---close to TLS level---tree diameter measurements than can be further used to support large-area forest inventories. The bias-compensated RMSE of \qtyrange[range-phrase = --]{5}{10}{\milli \metre} is close to the accuracy limit achievable by approximating stems as circles, due to the bark roughness and out-of-roundness of the stems. Importantly, the method also holds potential to enable accurate diameter measurements using emerging low-cost sensors that typically  suffer from increased diameter bias caused by large beams and high noise level.

\section*{CRediT authorship contribution statement}
\label{sec: author contributions}

J. Muhojoki: Conceptualization, Methodology, Formal analysis, Software, Investigation, Data Curation, Writing - Original Draft, Visualization; N. Ervasti: Investigation, Writing - Review \& Editing; E. Hyyppä: Software, Writing - Review \& Editing; A. Erkkilä: Investigation, Writing - Review \& Editing; E. Honkanen: Investigation, Writing - Review \& Editing; T. Yrttimaa: Investigation, Supervision, Writing - Review \& Editing; M. Vastaranta: Resources, Funding acquisition, Supervision, Writing - Review \& Editing; J. Hyyppä: Writing - Review \& Editing, Supervision, Project administration, Funding acquisition.

\section*{Funding information}
\label{sec: funding information}
This study was supported by the Research Council of Finland (grant numbers 346382, 359203, 373290, 374285, 357906, 371790, 346383, 373288, 376270)

\section*{Data availability}
MLS and TLS data is available from the corresponding author upon request and will be later published in \url{https://www.scanforest.fi/data/}

\section*{Declaration of competing interest}
The authors declare that they have no known competing financial interests or personal relationships that could have appeared to influence the work reported in this paper.

\scriptsize{\bibliography{references}}

\normalsize
\begin{appendices}
\counterwithin{figure}{section}
\counterwithin{table}{section}
\onecolumn

\section{Sensitivity to trigger threshold}
\label{app: trigger threshold}

\begin{figure}[htbp]
    \centering

    \begin{subfigure}{0.48\textwidth}
        \centering
        \caption{$\sigma_r = 4$ mm}
        \includegraphics[width=\linewidth]{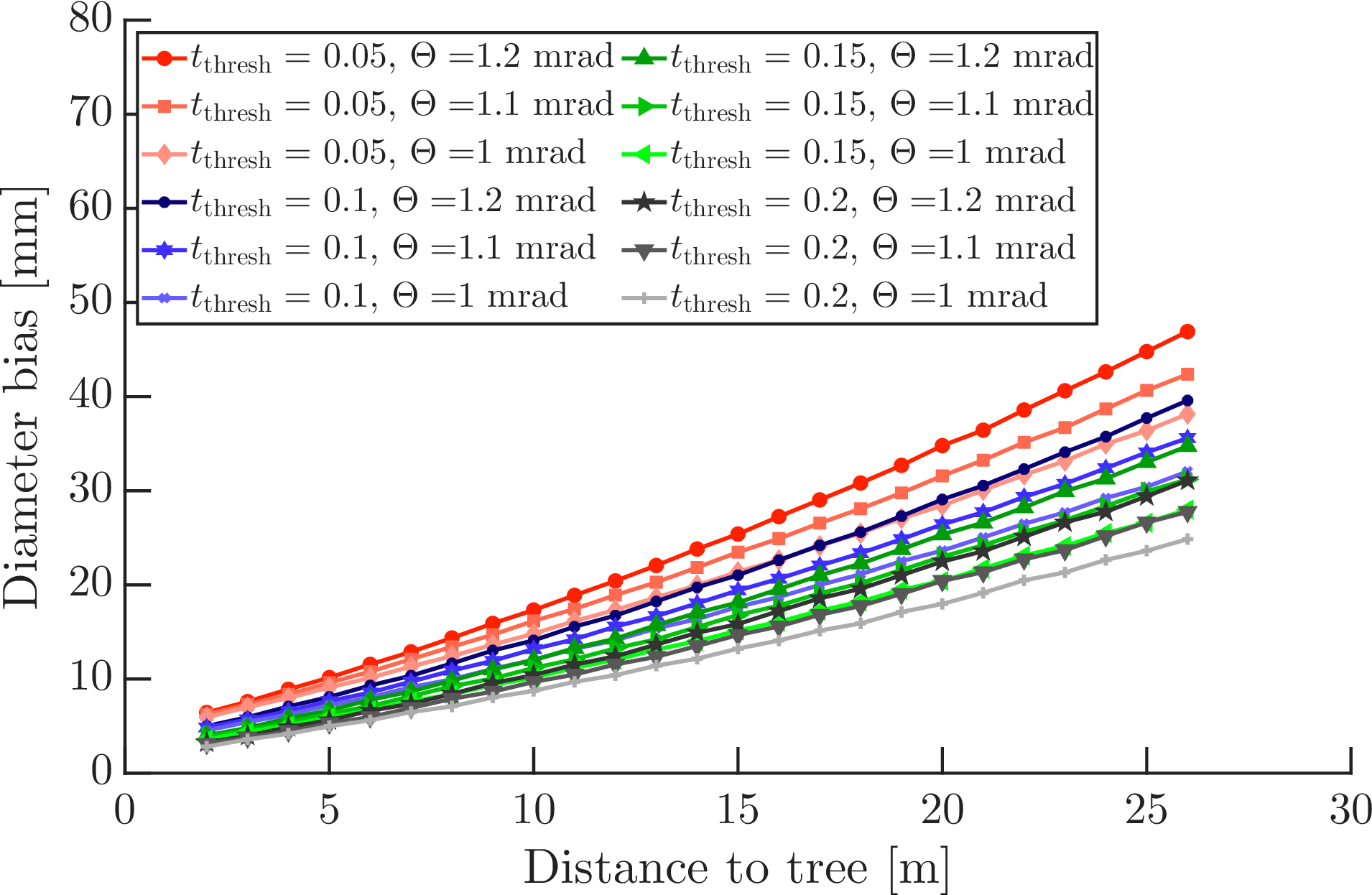}
    \end{subfigure}
    \hfill
    \begin{subfigure}{0.48\textwidth}
        \centering
        \caption{$\sigma_r = 6$ mm}
        \includegraphics[width=\linewidth]{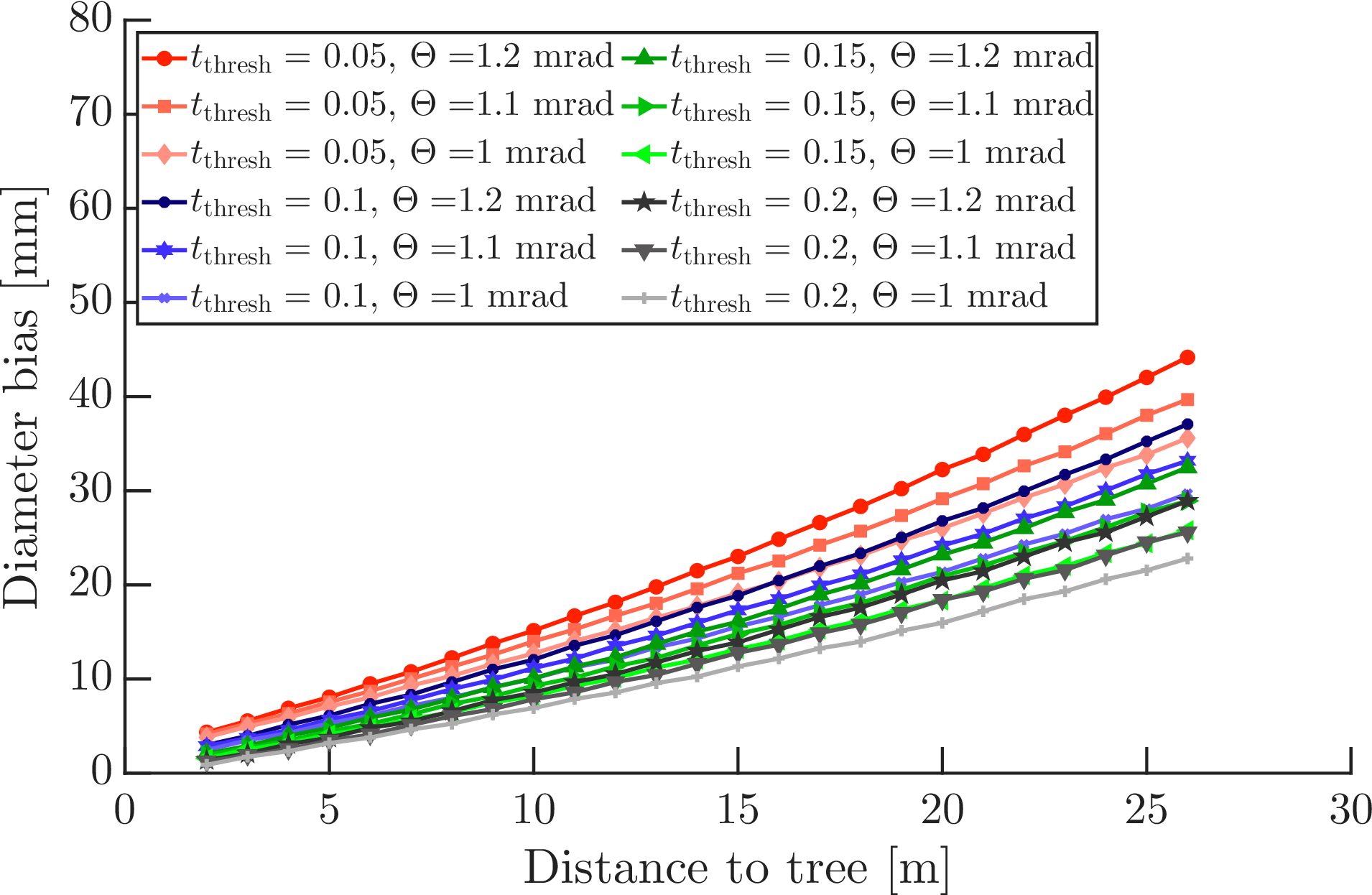}
    \end{subfigure}

    \vspace{0.5cm}

    \begin{subfigure}{0.48\textwidth}
        \centering
        \caption{$\sigma_r = 4$ mm}
        \includegraphics[width=\linewidth]{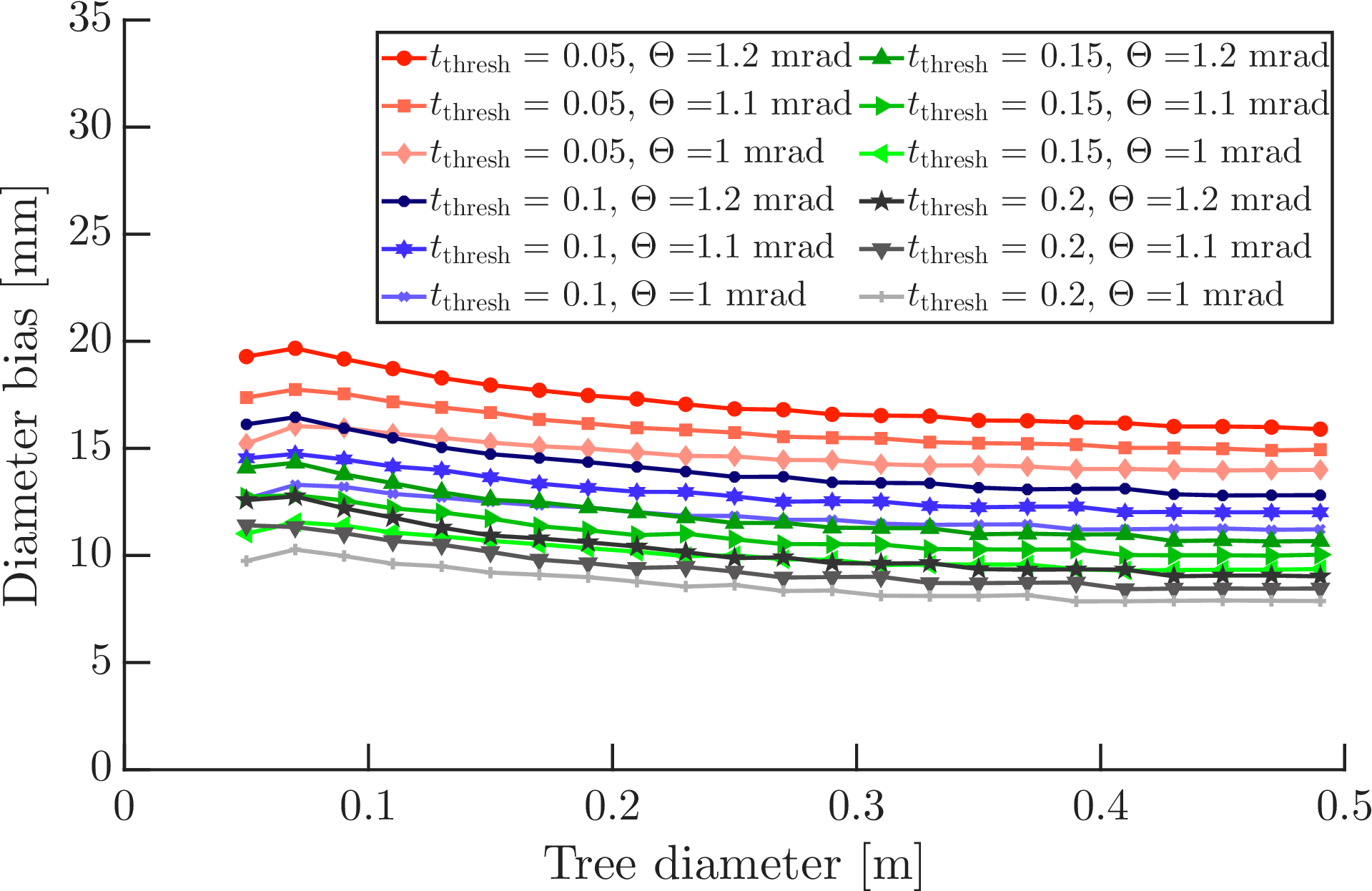}
    \end{subfigure}
    \hfill
    \begin{subfigure}{0.48\textwidth}
        \centering
        \caption{$\sigma_r = 6$ mm}
        \includegraphics[width=\linewidth]{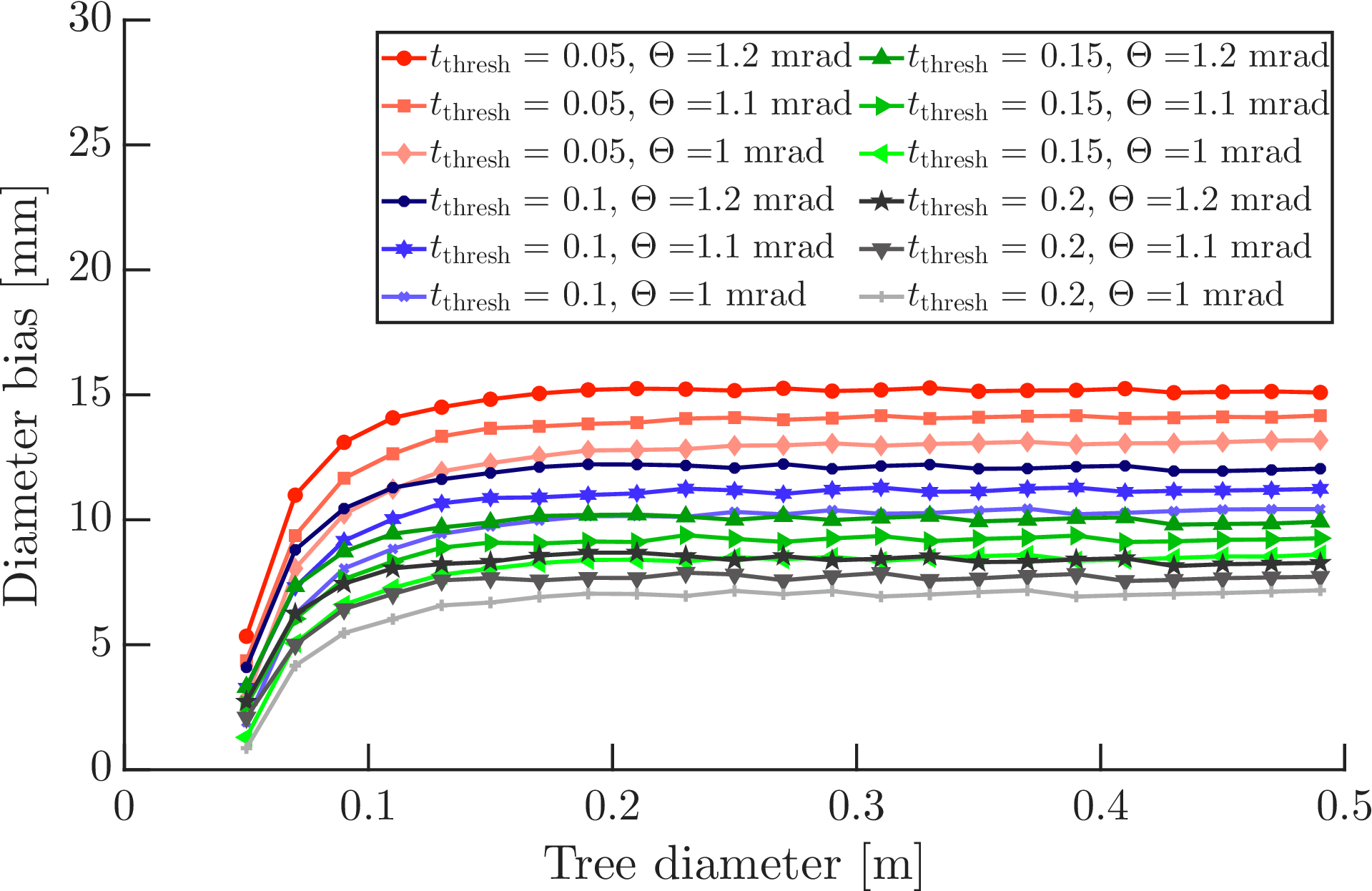}
    \end{subfigure}

    \vspace{0.5cm}

    \caption{Sensitivity analysis of the signal trigger threshold $t_\mathrm{thresh}$ Comparable to Figs.~\ref{fig: simulation results} and \ref{fig: simulation results large param spread}(e) and (f), where $t_\mathrm{thresh} = 0.1$. (a) and (b) present the tree diameter estimation bias as a function of measurement distance to the tree and  (c) and (d) as a function of tree diameter. In (a) and (c) $\sigma_r = \qty{4}{mm}$, and $\sigma_r = \qty{6}{mm}$ in (b) and (c). Colors are grouped based on the trigger threshold.}
    \label{fig: APPENDIX trigger threshold}
\end{figure}

Figure~\ref{fig: APPENDIX trigger threshold} presents the simulated diameter bias as a function of measurement distance to the target tree and its diameter The default values were \qty{10}{\metre} for the measurement distance and \qty{20}{cm} for the diameter. The selected simulation parameters are close to what was used for Faro Orbis in the bias compensation method with $t_\mathrm{thresh}= \ $ \qtyrange[range-phrase = --]{5}{20}{\percent}, $\Theta=\ $\qtyrange[range-phrase = --]{5}{20}{\percent}, and $\sigma_r = \{4,6\}$ mm. At short measurement distances, the parameter selection affects the simulated bias only little. At measurement distance of \qty{26}{\metre} and $\sigma_r =\qty{4}{mm}$, the bias ranged from \qty{31}{mm} to \qty{47}{mm} ($\Delta=\qty{16}{mm}$) with $\Theta = 1.2$ mrad, and from \qty{25}{mm} to \qty{38}{mm} ($\Delta=\qty{13}{mm}$) with $\Theta = 1.0$ mrad. For $\sigma_r =\qty{6}{mm}$ the results are similar, but the bias level is approximately \qty{3}{mm} lower throughout the curve. The sensitivity to $t_\mathrm{thresh}$ was similar as a function of tree diameter, where the difference between the extreme values of $t_\mathrm{thresh}$ was approximately \qtyrange[range-phrase = --]{6}{7}{mm}, depending slightly on the beam divergence, except for small diameters with ranging precision of \qty{6}{mm}, where the differences were significantly lower.

\end{appendices}
\end{document}